%% file: main.tex
\documentclass[10pt,twocolumn,letterpaper]{article}

\usepackage[pagenumbers]{cvpr} % To force page numbers, e.g. for an arXiv version

\input{preamble}

\definecolor{cvprblue}{rgb}{0.21,0.49,0.74}
\usepackage[pagebackref,breaklinks,colorlinks,allcolors=cvprblue]{hyperref}

\def\paperID{****} % *** 3045 Enter the Paper ID here
\def\confName{CVPR}
\def\confYear{2026}

\input{00_title}

\begin{document}

\maketitle

\renewcommand{\thefootnote}{\fnsymbol{footnote}}
\footnotetext[2]{Corresponding Authors}

\input{01_abstract}    
\input{02_introduction}

\input{03_related_work}
\input{04_methods}
\input{05_experiments}
\input{06_conclusions}
{
  \small
  \bibliographystyle{ieeenat}
  \bibliography{main}
}
\input{07_appendix}


\end{document}

%% file: preamble.tex
\usepackage{multirow}
\usepackage{bm}
\usepackage{graphicx}
\usepackage[most]{tcolorbox}
\usepackage{amssymb}
\usepackage{bbding}
\usepackage{xspace}
\usepackage{booktabs}   % 定义 \aboverrulesep, \belowrulesep 等
\usepackage{arydshln}   % 定义 \cdashline 以及 \adl@draw 等内部命令
\usepackage[table]{xcolor}

\makeatletter
\def\adl@drawiv#1#2#3{%
        \hskip.5\tabcolsep
        \xleaders#3{#2.5\@tempdimb #1{1}#2.5\@tempdimb}%
                #2\z@ plus1fil minus1fil\relax
        \hskip.5\tabcolsep}
\newcommand{\cdashlinelr}[1]{%
  \noalign{\vskip\aboverulesep
           \global\let\@dashdrawstore\adl@draw
           \global\let\adl@draw\adl@drawiv}
  \cdashline{#1}
  \noalign{\global\let\adl@draw\@dashdrawstore
           \vskip\belowrulesep}}

\definecolor{nmgray}{RGB}{229,229,229}
\definecolor{slightbule}{HTML}{D9E8F9}
\definecolor{slightgreen}{HTML}{C5E0B4}
\definecolor{slightyellow}{HTML}{FBF2CE}
\definecolor{slightred}{HTML}{FBDEDC}
\definecolor{promptbg}{HTML}{BFD8EF}

\newtcbox{\promptkey}{on line,
  colback=promptbg,       % 背景色
  colframe=promptbg,      % 边框色（与背景同色=无边框效果）
  boxrule=0pt,            % 边框粗细
  arc=0pt,                % 圆角
  boxsep=0pt,             % 取消默认内边距，改用四向分别设置
  left=0.1em, right=0.1em,
  top=0.12ex, bottom=0.12ex
}

\newtcolorbox{promptbox}[1][slightbule]{
  width=\columnwidth,
  colback=#1!75!white, 
  colframe=#1!75!white, 
  boxsep=0pt, left=9pt, right=10pt, top=0.25em, bottom=0.25em,
  fontupper=\linespread{0.9}\selectfont
}

\renewcommand{\paragraph}[1]{\smallskip\noindent\textbf{#1.}} % \smallskip
\renewcommand{\subparagraph}[1]{\smallskip\noindent\textbf{\underline{#1.}}} % 

\renewcommand{\sec}[1]{\textbf{\S#1}}

\newcommand{\tb}[1]{\textbf{#1}}
\newcommand{\ul}[1]{\underline{#1}}
\newcommand{\algo}[1]{\textbf{#1}\xspace}

\definecolor{R_blue}{HTML}{0070C0}
\definecolor{R_green}{HTML}{be9000}
\definecolor{R_red}{HTML}{c00000}

%% file: 00_title.tex
\title{Cut to the Chase: Training-free Multimodal Summarization via Chain-of-Events}
\author{
  Xiaoxing~You\textsuperscript{\rm 1} \quad
  Qiang~Huang\textsuperscript{\rm 2,}\footnotemark[2] \quad
  Lingyu~Li\textsuperscript{\rm 1} \quad
  Xiaojun~Chang\textsuperscript{\rm 3} \quad
  Jun~Yu\textsuperscript{\rm 2,}\footnotemark[2]
  \and  
  \textsuperscript{\rm 1}School of Computer Science, Hangzhou Dianzi University \\
  \textsuperscript{\rm 2}School of Intelligence Science and Engineering, Harbin Institute of Technology (Shenzhen) \\
  \textsuperscript{\rm 3}School of Information Science and Technology, University of Science and Technology of China 
  \and
  \tt\small \{youxiaoxing,~lilingyu0571\}@hdu.edu.cn, \{huangqiang,~yujun\}@hit.edu.cn, xjchang@ustc.edu.cn \\
}

%% file: 01_abstract.tex
\begin{abstract}
Multimodal Summarization (MMS) aims to generate concise textual summaries by understanding and integrating information across videos, transcripts, and images.
%%% challenges - more specific
However, existing approaches still suffer from three main challenges: (1) reliance on domain-specific supervision, (2) implicit fusion with weak cross-modal grounding, and (3) flat temporal modeling without event transitions.
%%% our solution - stronger positioning
To address these issues, we introduce \algo{CoE}, a \emph{training-free} MMS framework that performs structured reasoning through a \textbf{Chain-of-Events} guided by a Hierarchical Event Graph (HEG).
%%%
The HEG encodes textual semantics into an explicit event hierarchy that scaffolds cross-modal grounding and temporal reasoning.
Guided by this structure, \algo{CoE} localizes key visual cues, models event evolution and causal transitions, and refines outputs via lightweight style adaptation for domain alignment.
%%%
Extensive experiments on eight diverse datasets demonstrate that \algo{CoE} consistently outperforms state-of-the-art video CoT baselines, achieving average gains of \textbf{+3.04 ROUGE}, \textbf{+9.51 CIDEr}, and \textbf{+1.88 BERTScore}, highlighting its robustness, interpretability, and cross-domain generalization.
%%%
Our code is available at \url{https://github.com/youxiaoxing/CoE}.
\end{abstract}

%%% Multimodal Summarization (MMS) aims to generate concise textual summaries by understanding and integrating information across videos, transcripts, and images. However, existing approaches still suffer from three main challenges: (1) reliance on domain-specific supervision, (2) implicit fusion with weak cross-modal grounding, and (3) flat temporal modeling without event transitions. To address these issues, we introduce **CoE**, a training-free MMS framework that performs structured reasoning through a **Chain-of-Events** guided by a Hierarchical Event Graph (HEG). The HEG encodes textual semantics into an explicit event hierarchy that scaffolds cross-modal grounding and temporal reasoning. Guided by this structure, **CoE** localizes key visual cues, models event evolution and causal transitions, and refines outputs via lightweight style adaptation for domain alignment. Extensive experiments on eight diverse datasets demonstrate that **CoE** consistently outperforms state-of-the-art video CoT baselines, achieving average gains of **+3.04 ROUGE**, **+9.51 CIDEr**, and **+1.88 BERTScore**, highlighting its robustness, interpretability, and cross-domain generalization. Our code is available at https://github.com/youxiaoxing/CoE.

%% file: 02_introduction.tex
\section{Introduction}
\label{sec:intro}

\begin{figure*}[t]
  \centering
  \includegraphics[width=0.99\textwidth]{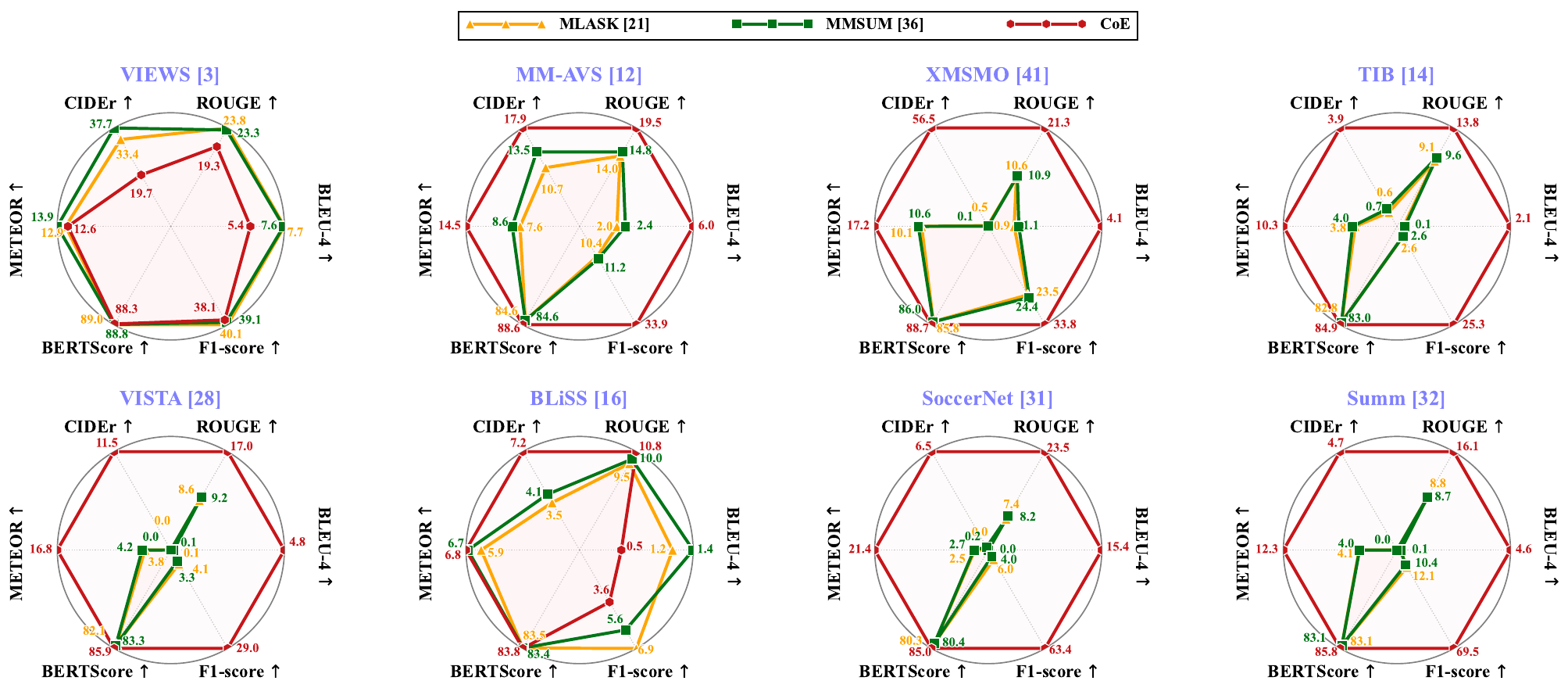}
  \vspace{-0.5em}
  \caption{\textbf{Motivating Experiments.} Existing MMS models (e.g., \algo{MLASK}~\cite{krubinski2023mlask} and \algo{MMSum}~\cite{qiu2024mmsum}) achieve strong in-domain results when trained on VIEWS~\cite{ayyubi2024views}, but their performance drops sharply under domain shift. In contrast, our \textbf{training-free} \algo{CoE} framework generalizes effectively across diverse datasets, maintaining stable zero-shot performance without task-specific training or adaptation.}
  \label{fig:motivation}
  \vspace{-0.5em}
\end{figure*}

%%% problem definition
Multimodal Summarization (MMS) aims to produce concise textual summaries from multi-source inputs such as videos, transcripts, and images.
In this work, we focus on the setting where the input is a video paired with its article or transcript, and the goal is to generate a compact, coherent textual summary.
%%% applications
Such summarization enables efficient comprehension and retrieval across diverse domains, including instructional videos~\cite{sanabria18how2, he2023bliss, beedu2025hiersum}, lectures~\cite{liu2025vista, gigant2023tib, tan2025smsmo}, and news broadcasts~\cite{li2020vmsmo, qiu2024mmsum, fu2021mm-avs, tang2023tldw}.

%%% existing work  
Most existing MMS methods adopt an \emph{encoder-fusion-decoder} framework~\cite{li2020vmsmo, fu2021mm-avs, krubinski2023mlask, tang2023tldw, qiu2024mmsum, he2023bliss, liu2024sitransformer}, where video and text are encoded separately, fused through cross-modal attention, and decoded into summaries.
%%%
Representative approaches such as \algo{MLASK} \cite{krubinski2023mlask} and \algo{MMSum} \cite{qiu2024mmsum} exemplify this paradigm but remain heavily reliant on task-specific supervision and fine-tuning.

%%% MLLM
Multimodal Large Language Models (MLLMs)~\cite{bai2025qwen2, hurst2024gpt, comanici2025gemini} have recently enhanced video understanding, inspiring efforts to extend the \emph{Chain-of-Thought (CoT)} reasoning paradigm~\cite{wei2022chain, wu2023role, wang2023plan, jin2024graphcot} to video.
%%%
These works range from building human-annotated video CoT datasets~\cite{wang2024videocot, han2025videoespresso, lee2025video-skill, cheng2025vstar} to developing grounded CoT models that align reasoning steps with visual evidence~\cite{fei2024vot, ghazanfari2025cof, zhang2025vitcot, liao2024videoinsta}. 
%%%
Efficiency-aware variants~\cite{arnab2025tcot, hu2025cos, wang2025videochat-a1, zhang2026silvr} further compress or selectively process frames for scalable reasoning.
%%%
Although primarily designed for Video Question Answering (VideoQA), these methods share a video-to-text reasoning pipeline similar to MMS and thus serve as valuable baselines.

Despite these advances, existing approaches still suffer from three fundamental limitations:
\begin{itemize}[nolistsep]
  \item \textbf{Reliance on Domain-specific Supervision:} 
  Existing MMS models rely heavily on large paired datasets and domain-specific fine-tuning~\cite{ali2025systematic}, hindering scalability and generalization to unseen domains. 
  %%%
  To examine this issue, we perform a domain generalization experiment (Figure~\ref{fig:motivation}), where two strong baselines, \algo{MLASK}~\cite{krubinski2023mlask} and \algo{MMSum}~\cite{qiu2024mmsum}, are trained on VIEWS~\cite{ayyubi2024views} and evaluated on out-of-domain datasets, revealing significant performance degradation. Comprehensive results are presented in Appendix~\ref{sec:appen:general}.
  %%%
  This reliance on annotated supervision hinders adaptation to low-resource or emerging domains where large-scale video-text pairs are unavailable.

  \item \textbf{Implicit Fusion with Weak Cross-modal Grounding:} 
  Although recent work has improved alignment through attention or contrastive objectives, most frameworks still perform implicit fusion in latent space without explicit reasoning over visual-textual correspondences~\cite{ali2025systematic, qiu2024mmsum, liu2024sitransformer, krubinski2023mlask, li2020vmsmo}. 
  %%%
  This leads to fragile cross-modal grounding and occasional semantic drift between modalities.

  \item \textbf{Flat Temporal Modeling without Event Transitions:} 
  Despite notable progress in video understanding~\cite{ghazanfari2025cof, hu2025cos, zhang2025vitcot, arnab2025tcot} and strong performance on benchmarks such as ActivityNet-QA~\cite{yu2019activitynet}, VideoMME~\cite{fu2025videomme}, LVBench~\cite{wang2025lvbench}, and MVBench~\cite{li2024mvbench}, most video CoT models still treat videos as flat sequences of frames or clips.
  %%%
  Lacking explicit modeling of hierarchical events and causal transitions, they capture only local temporal patterns. 
  %%%
  Thus, while effective for short-horizon reasoning, these models struggle to maintain global event evolution and narrative coherence required for long-video summarization.
\end{itemize}

To overcome these limitations, we advocate \textbf{explicit hierarchical event modeling} rather than implicit holistic fusion.
%%%
Accordingly, we introduce \algo{CoE (Chain-of-Events)}, a \emph{training-free} reasoning framework that constructs a \emph{Hierarchical Event Graph (HEG)} from joint video-text inputs to capture event semantics, temporal dependencies, and multimodal correspondences.
%%%
By replacing latent fusion with structured reasoning, \algo{CoE} enables accurate, interpretable, and domain-adaptive summarization without fine-tuning.
%%%
Our main contributions are as follows:
\begin{itemize}
  \item \textbf{Training-free and Domain-adaptive Summarization:} 
  \algo{CoE} introduces a fully \emph{training-free} reasoning framework with a lightweight \emph{style adaptation} module that adjusts linguistic tone and formality across diverse domains, ensuring robustness without domain-specific supervision.

  \item \textbf{Hierarchical Cross-modal Grounding:} 
  A \emph{Hierarchical Event Graph (HEG)} explicitly encodes textual semantics into event hierarchies and aligns them with corresponding video segments and entity-relation triples for fine-grained, interpretable grounding.
  
  \item \textbf{Event-evolution Reasoning for Temporal Coherence:} 
  An \emph{event-evolution reasoning} module traces how sub-events emerge, persist, and transition, capturing causal and temporal dependencies that yield coherent long-horizon summarization.
\end{itemize}

We conduct comprehensive experiments on eight multimodal summarization datasets. 
%%%
Across these benchmarks, \algo{CoE} consistently outperforms state-of-the-art video CoT baselines on both lexical and semantic metrics, achieving average gains of \textbf{+3.04 ROUGE}, \textbf{+9.51 CIDEr}, and \textbf{+1.88 BERTScore}, reflecting stronger lexical and semantic alignment with human references.
%%%
Further ablation studies confirm that each \algo{CoE} module makes meaningful contributions to the overall performance gains, validating the effectiveness of its hierarchical reasoning design.

%% file: 03_related_work.tex
\section{Related Work}
\label{sec:related_work}

\begin{figure*}[t]
  \centering
  \includegraphics[width=0.99\textwidth]{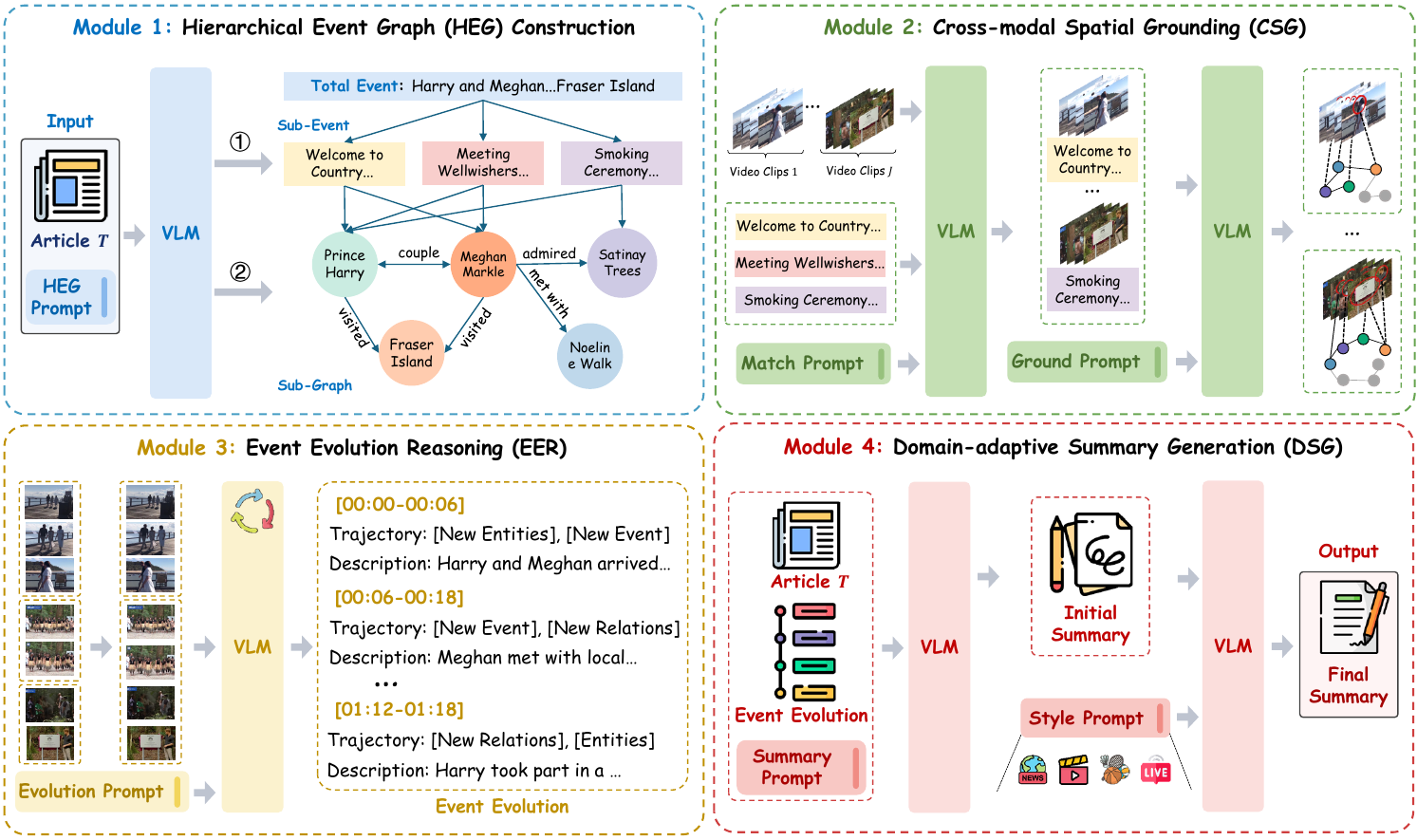}
  \vspace{-0.5em}
  \caption{\tb{Framework.} 
  \algo{CoE} is a training-free CoT framework for MMS with text-only output. 
  Given a video-text pair $(\bm{V}, \bm{T})$, Module~1 constructs a Hierarchical Event Graph (HEG) that organizes global events, sub-events, and entity-relation structures. 
  Guided by the HEG, Module~2 grounds video clips to sub-events and their visual entity-relation graphs. 
  Module~3 models event evolution by aggregating temporally coherent clips, and finally, Module~4 generates a domain-adaptive summary $\hat{s}$ from the resulting event trajectories.}
  \label{fig:framework}
  \vspace{-0.75em}
\end{figure*}

\paragraph{Multimodal Summarization (MMS)}
MMS aims to generate coherent summaries by integrating heterogeneous inputs such as videos, transcripts, and images.
%%% early works
Prior studies have explored MMS across diverse domains, including instructional and news videos, meeting recordings, and TV episodes, as well as multilingual and low-resource settings~\cite{sanabria18how2, fu2021mm-avs, gigant2023tib, papalampidi2023hierarchical3d, faheem2024urdumasd, liu2024multimodal}.
%%% recent advances
Recent advances extend MMS from text-only to multimodal outputs, generating both textual and visual summaries~\cite{li2020vmsmo, qiu2023sccs, tang2023tldw, he2023bliss, tan2025smsmo, liu2024sitransformer, beedu2025hiersum}.
%%%
Representative frameworks such as \algo{MLASK}~\cite{krubinski2023mlask} and \algo{MMSum}~\cite{qiu2024mmsum} exemplify supervised fusion-based architectures, while others explore diverse strategies:
\algo{SCCS}~\cite{qiu2023sccs} for segment-level alignment via optimal transport, \algo{SITransformer}~\cite{liu2024sitransformer} for filtering irrelevant modalities,
\algo{Uni-SciSum}~\cite{tan2025smsmo} for scientific multimodal data, and \algo{HierSum}~\cite{beedu2025hiersum} for hierarchical context modeling.

Despite progress, most MMS models remain \emph{training-intensive} and \emph{domain-specific}, requiring extensive supervision and domain-wise fine-tuning, which limits scalability and generalization.
In contrast, \algo{CoE} offers a \emph{training-free} MMS framework that enables generalizable multimodal reasoning without additional adaptation.

\paragraph{Video Chain-of-Thought (CoT)}
The CoT paradigm, initially developed for textual reasoning, has recently been extended to video understanding.
%%%
Several studies focus on data and benchmarks, developing reasoning-oriented video datasets to advance spatio-temporal cognition~\cite{wang2024videocot,han2025videoespresso,cheng2025vstar} and skill-aware reasoning~\cite{lee2025video-skill}. 
From the modeling perspective, video CoT methods explore visual grounding and efficient reasoning through adaptive frame selection, compression, and structured decomposition \cite{fei2024vot, liao2024videoinsta, wang2025videochat-a1, wang2025cotasks, tang2025cardiff, zhang2026silvr}. 

Among them, four representative approaches are closely aligned with the MMS setting:
\algo{ViTCoT}~\cite{zhang2025vitcot} interleaves video and text in the reasoning process, enriching textual reasoning with visual context;
\algo{CoF}~\cite{ghazanfari2025cof} grounds reasoning steps to discrete frame IDs, improving temporal interpretability;
\algo{TCoT}~\cite{arnab2025tcot} iteratively selects question-relevant frames to support long-video reasoning under limited context; and
\algo{CoS}~\cite{hu2025cos} performs contrastive shot-level reasoning to highlight salient content.

%%% comparison
While these models excel at question-conditioned, local reasoning, they struggle to capture global event evolution and long-range narratives.
In contrast, \algo{CoE} adopts a global event-centric perspective, constructing a Hierarchical Event Graph (HEG) from joint video-text inputs to model event hierarchies, causal transitions, and temporal dependencies for coherent summarization.

%% file: 04_methods.tex
\section{The CoE Framework}
\label{sec:method}

In this work, we focus on \tb{Multimodal Summarization (MMS) with text-only output}, where the input consists of a video $\bm{V}$ and its accompanying text $\bm{T}$, and the goal is to produce a textual summary $\hat{s}$:
\begin{equation} \label{eq:definition}
\hat{s} = f(\bm{V}, \bm{T}),
\end{equation}
where $f(\cdot)$ denotes the MMS model, $\bm{V} =(v_{1}, \cdots, v_{N})$ is the sequence of video frames, and $\bm{T} = (t_{1}, \cdots, t_{M})$ is the textual input, where each $t_i$ corresponds to a token.

\subsection{Overview}
\label{sec:method:overview}

We introduce \algo{CoE}, a training-free framework that performs structured multimodal reasoning through a Hierarchical Event Graph (HEG).
As depicted in Figure \ref{fig:framework}, \algo{CoE} decomposes MMS into four interconnected modules:
\begin{enumerate}[nolistsep,label*=(\arabic*)]
  \item \tb{Hierarchical Event Graph (HEG) Construction (\sec{\ref{sec:method:hierarchical-event}}):} 
  Converts input text into a three-level HEG that organizes global events, sub-events, and entity-relation structures as the semantic scaffold for reasoning.

  \item \tb{Cross-modal Spatial Grounding (CSG) (\sec{\ref{sec:method:spatial-grounding}}):} 
  Uses the HEG to guide video interpretation, aligning clips with sub-event anchors and grounding entity-relation triples to produce visually supported subgraphs.

  \item \tb{Event Evolution Reasoning (EER) (\sec{\ref{sec:method:event-evolution}}):} 
  Merges semantically coherent clips into temporal segments and analyzes changes in their subgraphs to derive event trajectories that describe how entities and relations evolve.

  \item \tb{Domain-adaptive Summary Generation (DSG) (\sec{\ref{sec:method:summary-geneation}}):} 
  Synthesizes these trajectories into an initial summary and refines it via lightweight style adaptation to align with domain-specific linguistic conventions.
\end{enumerate}

Together, these modules unify event-level representation and spatio-temporal reasoning into a training-free pipeline for robust and generalizable multimodal summarization.

\subsection{Hierarchical Event Graph (HEG) Construction}
\label{sec:method:hierarchical-event}

\paragraph{Hierarchical Semantic Representation}
To structure event-rich narratives, \algo{CoE} constructs an \tb{HEG} that encodes the input text and provides contextual grounding for subsequent reasoning.
%%%
As described in Figure \ref{fig:framework}(a), the HEG consists of three layers:
\begin{itemize}
  \item \tb{Global Event Layer:}~Captures the overall theme of the narrative (e.g., the main story in a news article).
  
  \item \tb{Sub-Event Layer:}~Decomposes the global event into coherent components (e.g., \textit{earthquake occurrence}, \textit{emergency rescue}, and \textit{post-disaster reconstruction} in an \textit{earthquake relief} report).

  \item \tb{Entity-Relation Layer:}~Models the key entities and their interactions within sub-events. Shared entities connect sub-events into a unified, hierarchical graph.
\end{itemize}

HEG serves as a \tb{semantic backbone} for zero-shot multimodal reasoning and summary generation.

\paragraph{HEG Construction}
To build the HEG, \algo{CoE} first identifies the global event $g$ and then decomposes it into sub-events $\mathcal{H} = \{h_{k}\}_{k=1}^{K}$, where $K$ denotes the number of sub-events:
\begin{promptbox}
Given [$\bm{T}$], identify the high-level [$g$] that summarizes the overall topic, then decompose [$g$] into a set of [$\mathcal{H}$] that represent distinct aspects of the main event.
\end{promptbox}

Then, for each sub-event $h_{k}$, \algo{CoE} extracts the entity set $\mathcal{E}_k = \{e_q\}_{q=1}^{Q}$ and relations $\mathcal{R}_k = \{r_{l}\}_{l=1}^{L}$ to form a subgraph $\mathcal{G}_k = (\mathcal{E}_k, \mathcal{R}_k)$:
\begin{promptbox}
Given [$h_k$] and [$\bm{T}$], extract all relevant [$\mathcal{E}_k$] and their explicitly stated [$\mathcal{R}_k$], and construct a [$\mathcal{G}_k$].
\end{promptbox}

To improve the robustness of LLM-based extraction, we constrain the prompts and restrict the output space to ensure structural consistency (details in Appendix~\ref{sec:appen:prompt}).

\subsection{Cross-modal Spatial Grounding (CSG)}
\label{sec:method:spatial-grounding}

While the HEG offers a structured semantic scaffold, effective multimodal summarization also requires grounding these textual structures in visual evidence.
%%%
Long videos often contain overlapping and temporally dispersed events, making direct frame-to-text alignment unreliable. 

To address this, \textbf{CSG} reasons over the video stream to locate visual segments that correspond to each sub-event and its associated entities, using HEG sub-event nodes as semantic anchors for interpreting video content.

\paragraph{Sub-Event Alignment}
Given a video $\bm{V}$, we uniformly sample $N$ frames and segment them into short clips $\{\bm{C}_j\}_{j=1}^{J}$, each capturing a coherent local temporal context. 
Under the guidance of the HEG, each clip is aligned with the most relevant sub-event, where the HEG provides semantic context for event-level interpretation:
\begin{promptbox}[slightgreen]
Given [$\bm{C}_j$] and [$\mathcal{H}$], identify the [$h_k$] that best matches the clip using visual and textual cues.
\end{promptbox}

\paragraph{Entity-Relation Grounding}
After aligning clips to sub-events, the model identifies visually supported entities and their interactions within each clip.
%%%
The VLM describes the scene as \emph{entity-relation-entity} triples, constructing a visually grounded graph $\mathcal{G}_{k}^{(j)} = (\mathcal{E}_{k}^{(j)}, \mathcal{R}_{k}^{(j)})$ as a subset of the textual reference $\mathcal{G}_{k}$:
\begin{promptbox}[slightgreen]
Given [$\bm{C}_j$] and [$\mathcal{G}_{k}$], extract [$\mathcal{G}_{k}^{(j)}$] by identifying the entities and relations that are supported in this scene.
\end{promptbox}
%%%
This process enhances cross-modal grounding with relational semantics, yielding fine-grained correspondences between visual observations and textual concepts.

\subsection{Event Evolution Reasoning (EER)}
\label{sec:method:event-evolution}

To capture long-range temporal dependencies, \algo{CoE} applies \textbf{EER} on the clip-level grounded graphs from CSG. 
%%%
This module establishes temporal continuity between isolated clips, forming a coherent sequence of event transitions.

\begin{figure}[t]
  \centering
  \includegraphics[width=0.99\linewidth]{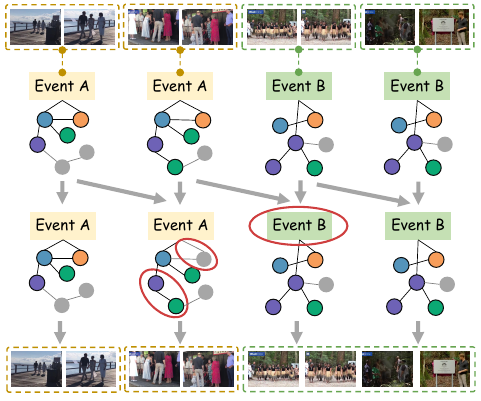}
  \vspace{-0.75em}
  \caption{\tb{Video Clip Aggregation.} 
  \algo{CoE} merges adjacent clips grounded to the same sub-event and sharing identical entity-relation graphs into longer temporal segments. Red circles indicate new entities or sub-event changes that trigger a new segment.}
  \label{fig:clip_merge}
  \vspace{-0.75em}
\end{figure}

\paragraph{Video Clip Aggregation}
Building on clip-level grounding (\sec{\ref{sec:method:spatial-grounding}}), \algo{CoE} merges semantically coherent clips into longer temporal segments (Figure~\ref{fig:clip_merge}). 
%%%
Adjacent clips $\bm{C}_{j}$ and $\bm{C}_{j+1}$ belonging to the same sub-event $h_{k}$ are merged when their subgraphs $\mathcal{G}_{k}^{(j)}$ and $\mathcal{G}_{k}^{(j+1)}$ are identical. 
%%%
Merging continues until a sub-event change, graph difference, or the segment-length limit is reached, producing merged segments $\bm{C}_{p} = \{\bm{C}_{j}, \bm{C}_{j+1}, \cdots, \bm{C}_{j'}\}$ with associated subgraphs $\mathcal{G}_{k}^{(p)}$.

\paragraph{Graph-based Evolution Analysis}
After temporal aggregation, \algo{CoE} analyzes how subgraphs evolve across segments to infer event dynamics.
%%%
For each segment $\bm{C}_{p}$, its subgraph $\mathcal{G}_{k}^{(p)}$ is compared with the previous one $\mathcal{G}_{k'}^{(p-1)}$ to detect emerging, persisting, or vanishing entities and relations.
These structural variations define \tb{event trajectories}, capturing how narrative context and interactions evolve. % over time.

Formally, for each $\bm{C}_{p}$, the VLM generates an event-trajectory description $\mathcal{D}_p$ that synthesizes visual evidence, temporal context, and relevant textual information:
\begin{promptbox}[slightyellow]
Given [$\bm{T}$], [$\bm{C}_{p}$], [$\mathcal{G}_{k}^{(p)}$], and  [$\mathcal{G}_{k'}^{(p-1)}$], analyze the event trajectory and generate a concise description [$\mathcal{D}_p$] of the key developments.
\end{promptbox}

Through \textbf{EER}, \algo{CoE} converts spatially grounded clip-level graphs into a sequence of temporally coherent event trajectories that form the basis for summary generation.

\subsection{Domain-adaptive Summary Generation (DSG)}
\label{sec:method:summary-geneation}

Finally, \textbf{DSG} synthesizes an event-centric summary and refines it to match domain-specific linguistic styles.

\paragraph{Event-centric Summary Generation}
Given the event-trajectory descriptions $\{\mathcal{D}_p\}_{p=1}^{P}$ produced by \textbf{EER}, \algo{CoE} composes an initial summary $\hat{s}_{\text{init}}$:
\begin{promptbox}[slightred]
Given the sequence of event trajectory descriptions [$\{\mathcal{D}_p\}_{p=1}^{P}$] derived from video analysis, synthesize a coherent and comprehensive summary [$\hat{s}_{\text{init}}$].
\end{promptbox}

This initial draft consolidates multimodal reasoning into a unified textual narrative, serving as the content-preserving base for stylistic refinement.

\paragraph{Style Adaptation}
Different domains follow distinct linguistic and rhetorical conventions.
%%%
To ensure domain alignment, \algo{CoE} leverages a small set of reference summaries $\mathcal{Y}_{\text{ref}} = \{y_{\text{ref}}^1, \cdots, y_{\text{ref}}^R\}$ sampled from the target domain, which serve as stylistic exemplars.
%%%
The VLM then adapts $\hat{s}_{\text{init}}$ to match the target domain’s expression while preserving factual accuracy and logical flow:
\begin{promptbox}[slightred]
Given [$\hat{s}_{\text{init}}$] and [$\mathcal{Y}_{\text{ref}}$], adapt [$\hat{s}_{\text{init}}$] to match the linguistic style, producing the final summary [$\hat{s}$].
\end{promptbox}

This process refines tone, phrasing, and discourse structure to match target-domain conventions while maintaining content fidelity, yielding the final summary $\hat{s}$.

%% file: 05_experiments.tex
\section{Experiments}
\label{sec:expt}

We comprehensively evaluate \algo{CoE} to address the following research questions (RQs):
\begin{itemize}
  \item \textbf{RQ1:} How effective is \algo{CoE} compared to state-of-the-art baselines across diverse datasets? (\sec{\ref{sec:expt:comparison}})
  \item \textbf{RQ2:} What are the contributions of individual modules within the framework? (\sec{\ref{sec:expt:ablation}})
  \item \textbf{RQ3:} Does \algo{CoE} maintain consistent performance across different MLLM backbones? (\sec{\ref{sec:expt:backbone}})
  \item \textbf{RQ4:} How does the parameter size of the MLLM affect the performance of the framework? (\sec{\ref{sec:expt:modelsize}})
  \item \textbf{RQ5:} Can \algo{CoE} produce domain-adaptive, well-grounded, and temporally coherent summaries? (\sec{\ref{sec:expt:case}})
  % How does the hyperparameter setting affect performance across different datasets? 
\end{itemize}

We first present the experimental setup, followed by results and analysis for each of the above research questions.

\subsection{Experimental Setup}
\label{sec:expt:setup}

\paragraph{Datasets}
We evaluate \algo{CoE} on eight multimodal summarization benchmarks spanning news, instructional, sports, and TV domains: \tb{VIEWS}~\cite{ayyubi2024views}, \tb{MM-AVS}~\cite{fu2021mm-avs}, XMSMO-News (\tb{XMSMO})~\cite{tang2023tldw}, \tb{TIB}~\cite{gigant2023tib}, \tb{VISTA}~\cite{liu2025vista}, \tb{BLiSS}~\cite{he2023bliss}, SoccerNet-Caption (\tb{SoccerNet})~\cite{mkhallati2023soccernet}, and SummScreen$^{\text{3D}}$ (\tb{Summ})~\cite{papalampidi2023hierarchical3d}. 
%%%
Dataset statistics and domain characteristics are detailed in Appendix~\ref{sec:appen:dataset}.
%%%
For datasets (VIEWS and VISTA) lacking textual annotations, we leverage video descriptions from YouTube for \textbf{VIEWS} and employ Whisper~\cite{radford2023whisper} to generate transcripts for \textbf{VISTA}.

\paragraph{Evaluation Metrics}
We adopt standard summarization metrics: \textbf{BLEU-4}~\cite{papineni2002bleu} and \textbf{ROUGE}~\cite{lin2004rouge} for lexical overlap,
\textbf{CIDEr}~\cite{vedantam2015cider} for reference consensus, and \textbf{METEOR}~\cite{denkowski2014meteor} and \textbf{BERTScore}~\cite{zhangbertscore} for semantic fidelity.
%%%
We further include an entity-level \textbf{F1-score} using spaCy~\cite{honnibal2017spacy} to evaluate the accuracy of named entities, reflecting factual grounding and referential consistency. 
%%%
Finally, we employ \textbf{G-Eval}~\cite{liu2023g} as an LLM-as-a-judge metric to assess generation quality, where the final score is computed by averaging the four dimension-wise ratings, and then averaging the scores produced by GPT-5.2 and Claude-Sonnet-4.5.

\paragraph{Baselines}
We compare \algo{CoE} with four state-of-the-art video CoT models: \algo{TCoT}~\cite{arnab2025tcot}, \algo{CoF}~\cite{ghazanfari2025cof}, \algo{ViTCoT}~\cite{zhang2025vitcot}, and \algo{CoS}~\cite{hu2025cos}, all of which perform step-wise video reasoning and produce text-only outputs.
%%%
All models are evaluated in a \emph{zero-shot, training-free} setting 
with standardized inference configurations (see Appendix~\ref{sec:appen:imple}).

\paragraph{Implementation Details}
Unless otherwise specified, \algo{CoE} employs Qwen2.5-VL-7B-Instruct~\cite{bai2025qwen2} as its default multimodal backbone. 
Up to 72 frames are uniformly sampled per video and divided into 12 clips, each containing up to 30 frames after temporal merging. We set $R{=}5$ and sample five summaries from the training set as $\mathcal{Y}_{\text{ref}}$.
%%% 
\algo{TCoT} and \algo{ViTCoT} are instantiated with the same backbone, with \algo{ViTCoT} leveraging CLIP~\cite{radford2021clip} for interleaved CoT representation~\cite{zhang2025vitcot}.
\algo{CoF}~\cite{ghazanfari2025cof} uses the official InternVL3-8B~\cite{zhu2025internvl3} checkpoint, while \algo{CoS}~\cite{hu2025cos} follows its original design with LLaVA-Video~\cite{zhang2024videollava} for binary encoding and LongVA~\cite{zhang2025long} for summary generation.
%%%
All baselines adhere to their released configurations, with details in Appendix~\ref{sec:appen:imple}.

\paragraph{Experiment Environment}
All experiments are conducted using VLLM~\cite{kwon2023vllm} and PyTorch~\cite{paszke2019pytorch} on four NVIDIA L20 GPUs (48 GB VRAM), ensuring consistent inference conditions across all models. The proprietary GPT-5 model is accessed via its official API.

\subsection{Comparison with Baselines}
\label{sec:expt:comparison}

\input{tables/main_exprimental}

Table~\ref{tab:main-results} compares \algo{CoE} with four advanced video CoT baselines on eight MMS datasets. \algo{CoE} achieves the best performance on most dataset-metric pairs, demonstrating consistent superiority across lexical, semantic, and factual dimensions. Inference efficiency is reported in Appendix~\ref{sec:appen:time}.

\paragraph{Lexical Overlap}
\algo{CoE} exhibits strong alignment with human references across all metrics.
%%%
On CIDEr, it ranks first on all eight datasets, averaging \textbf{+9.51} points, with a remarkable \textbf{+31.73} gain on XMSMO over the best baseline \algo{TCoT}.
%%%
For ROUGE, \algo{CoE} leads on seven datasets (e.g., \textbf{+9.56} on SoccerNet) and remains competitive on BLiSS.
%%%
It also achieves top or second-best results on BLEU-4 and METEOR, confirming that explicit event–entity modeling effectively captures salient lexical content compared with latent-fusion CoT methods.

\paragraph{Semantic Faithfulness}
On BERTScore, \algo{CoE} attains the highest results on seven datasets, with notable gains of \tb{+2.93}, \tb{+3.77}, and \tb{+3.43} on MM-AVS, SoccerNet, and Summ, respectively. 
%%%
Although slightly below the best baseline on BLiSS (--1.13 points), \algo{CoE} still excels on CIDEr and BLEU-4, showing strong overall semantic fidelity.

\paragraph{Entity Accuracy}
Measured by entity-level F1-score, \algo{CoE} outperforms all baselines on seven datasets, often by large margins.
%%%
On SoccerNet, \algo{CoE} reaches \textbf{63.37}, compared to \textbf{31.45} for the strongest baseline \algo{CoS}, corresponding to an absolute gain of \textbf{+31.92}. 
%%%
Similarly, on Summ, it achieves \textbf{69.50} versus \textbf{31.06} for \algo{CoS}, yielding an absolute improvement of \textbf{+38.44}. 
%%%
These results underscore the role of HEG-based entity tracking in maintaining referential coherence throughout long videos.

\paragraph{LLM-as-a-Judge Quality}
As shown in Table~\ref{tab:main-results}, \algo{CoE} achieves the highest G-Eval scores on six datasets and ranks second on the others.
%%%
It gains \tb{+0.70} G-Eval on MM-AVS and \tb{+1.03} on Summ, suggesting that explicit event reasoning improves holistic quality as judged by LLMs.

\paragraph{Cross-Domain Generalization}
\algo{CoE} exhibits stable zero-shot performance across diverse domains without any adaptation, ranking within the top two on nearly all metrics.
%%%
In contrast to the high variance of baselines, this consistency demonstrates that hierarchical event-entity reasoning offers a robust, domain-agnostic semantic foundation well suited for real-world MMS applications.

\input{tables/ablation_rouge}

\input{tables/ablation_cider}
% \input{tables/ablation_bertscore}

\subsection{Ablation Study}
\label{sec:expt:ablation}

To quantify each module's contribution, we perform ablations by removing one component at a time.
Tables \ref{tab:ablation-rouge} and \ref{tab:ablation-cider} report ROUGE and CIDEr scores, respectively, for the full \algo{CoE} framework and its four ablated variants.
\begin{itemize}
  \item \textbf{Impact of HEG:}
  Removing HEG (\tb{CoE -- HEG}) yields the largest CIDEr drops (e.g., \textbf{--12.46} on XMSMO and \textbf{--2.52} on TIB) and consistent ROUGE declines (e.g., \textbf{--2.85} on MM-AVS and \textbf{--1.46} on XMSMO), underscoring its role as the global event scaffold.

  \item \textbf{Impact of CSG:}
  Disabling CSG (\tb{CoE -- CSG}) reduces CIDEr by 3.30 on average and ROUGE by 0.83, notably \textbf{--1.18} on MM-AVS and \textbf{--0.72} on VIEWS, confirming the importance of fine-grained entity–relation grounding.

  \item \textbf{Impact of EER:} 
  Removing EER (\tb{CoE -- EER}) retains basic structure but weakens temporal consistency, lowering CIDEr on seven datasets and ROUGE throughout, highlighting its role in modeling event trajectories, especially in long-horizon scenarios.

  \item \textbf{Impact of DSG}
  Without DSG (\tb{CoE -- DSG}), ROUGE changes slightly but CIDEr drops sharply (e.g., \textbf{--10.50} on VIEWS, \textbf{--15.99} on XMSMO), showing the impact of domain-specific style alignment.
\end{itemize}

Overall, \algo{CoE}'s performance stems from the synergy of its modules: \textbf{HEG} structures events, \textbf{CSG} grounds visual evidence, \textbf{EER} ensures temporal coherence, and \textbf{DSG} aligns stylistic conventions, together enabling semantically faithful and domain-adaptive summaries.

\begin{figure}[t]
  \centering
  \includegraphics[width=0.99\linewidth]{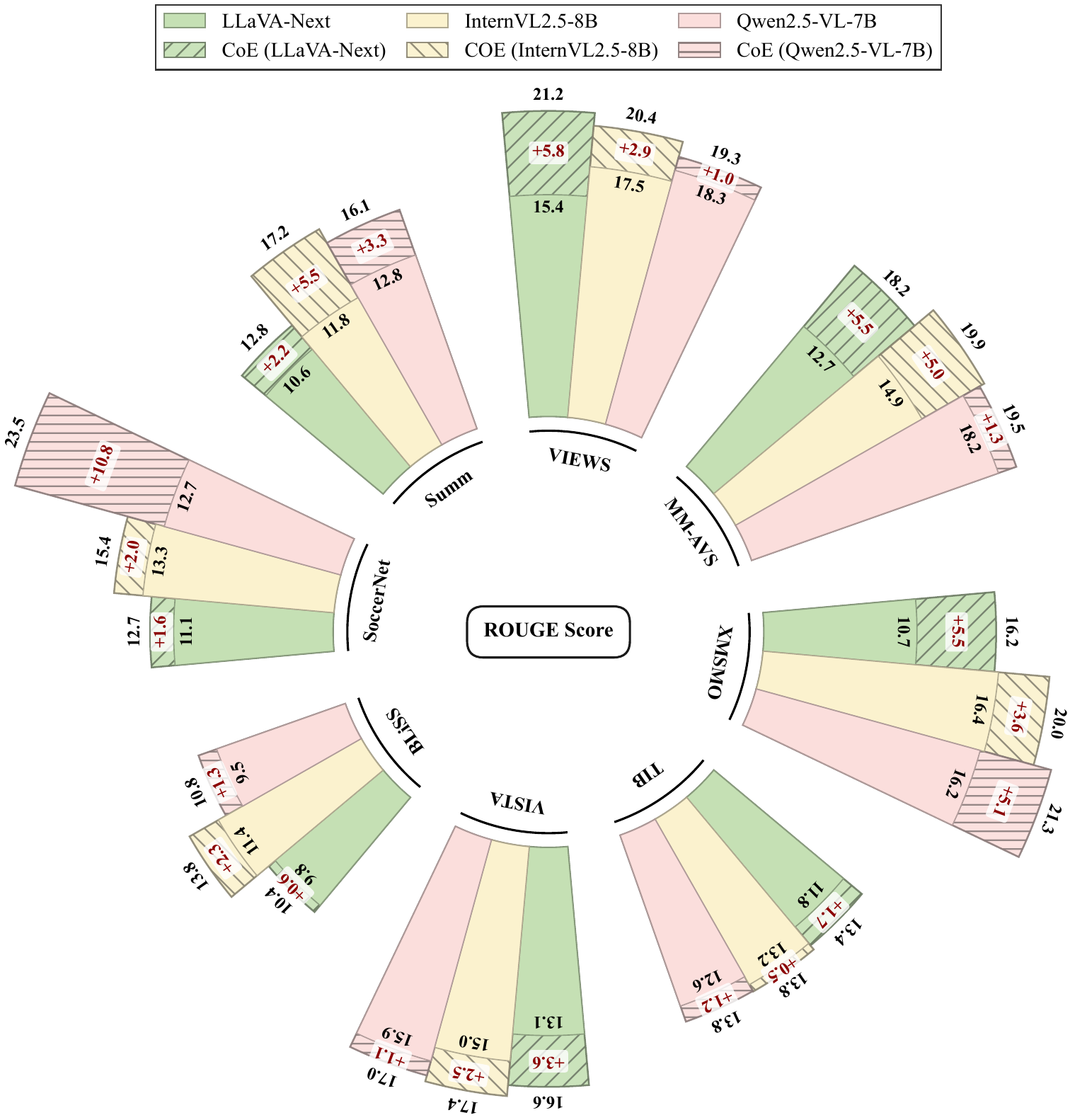}
  \vspace{-0.5em}
  \caption{\textbf{Backbone Generalization.} Performance across different backbones in terms of ROUGE score on eight datasets. For each backbone, we report the vanilla model and its CoE-based variant, denoted as \texttt{CoE(\,·\,)}. The consistent gains of \texttt{CoE(\,·\,)} over the corresponding vanilla backbones demonstrate the strong generalization ability and robustness of our approach across diverse architectures.}
  \label{fig:backbone-result}
  % \vspace{-0.5em}
\end{figure}

\begin{figure}[t]
  \centering
  \includegraphics[width=0.99\linewidth]{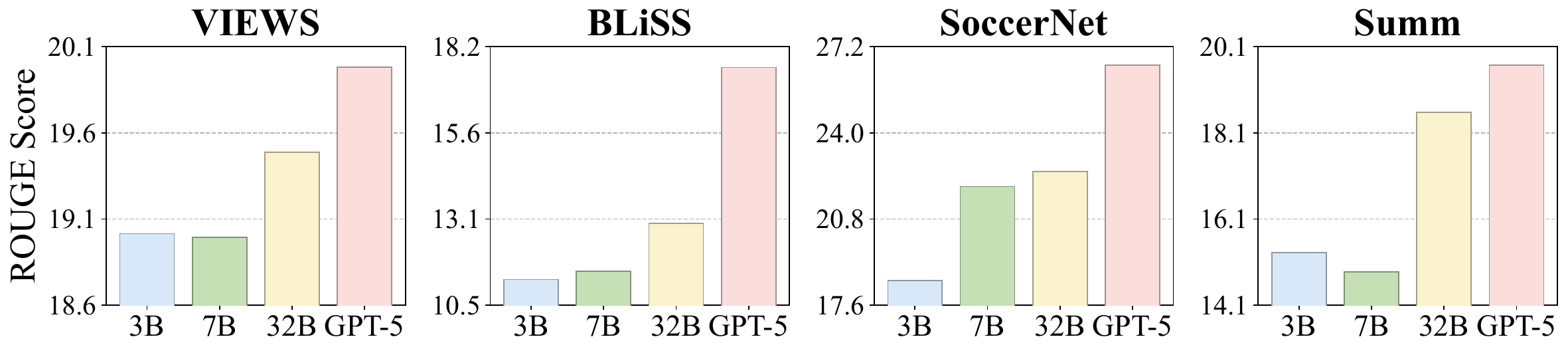}
  \vspace{-0.5em}
  \caption{\tb{Effect of Model Size.} Performance of \algo{CoE} with different MLLM backbones. We report results for Qwen2.5-VL models of increasing size (3B, 7B, 32B) and a proprietary GPT-5 model, showing a clear performance gain as model capacity grows across the four evaluation datasets.}
  \label{fig:model_size}
  % \vspace{-0.5em}
\end{figure}

\begin{figure*}[t]
  \centering
  \begin{subfigure}[t]{0.495\textwidth}
    \centering
    \includegraphics[width=\linewidth]{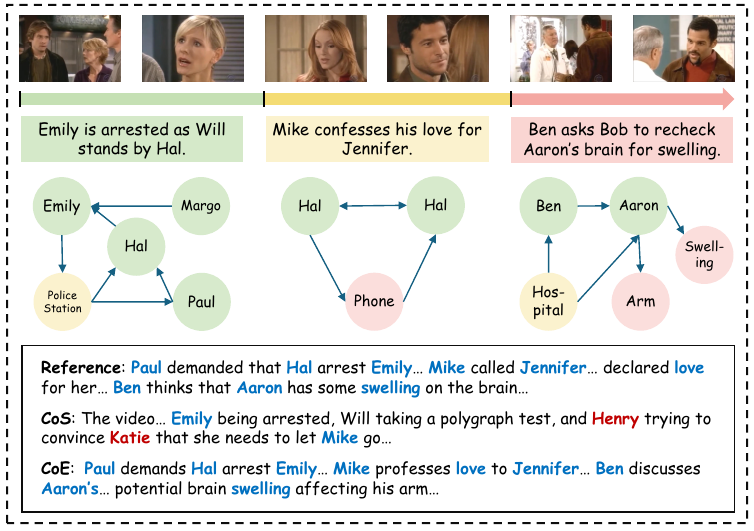}
    \caption{\algo{CoE} enables precise sub-event and entity grounding through HEG.}
    \label{fig:case_study_a}
  \end{subfigure}\hfill
  \begin{subfigure}[t]{0.495\textwidth}
    \centering
    \includegraphics[width=\linewidth]{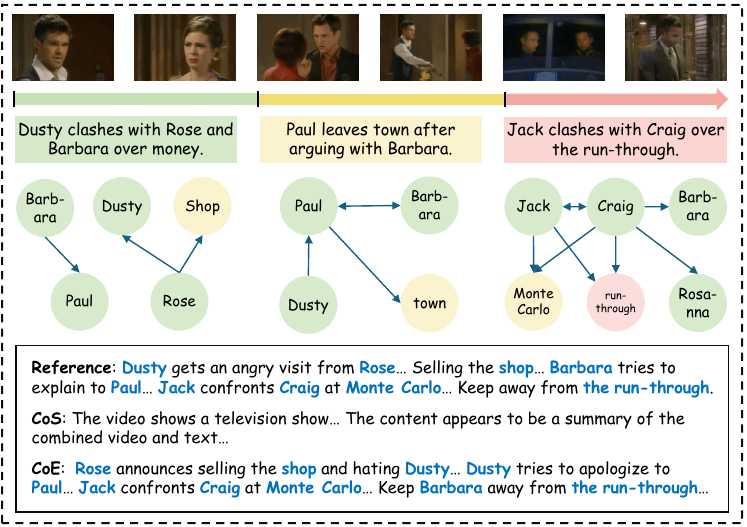}
    \caption{\algo{CoE} models event transitions and script style.}
    \label{fig:case_study_b}
  \end{subfigure}
  \vspace{-0.25em}
  \caption{\tb{Case Study}. 
  (a) Guided by the HEG, \algo{CoE} accurately assigns sub-events to segments and grounds both visible and script-referenced entities at a fine-grained level.
(b) \algo{CoE} captures event evolution and adapts to the TV-script style, whereas the baseline mainly describes local scenes, missing long-range transitions and stylistic alignment.}
  \vspace{-0.5em}
\end{figure*}

\subsection{Backbone Versatility}
\label{sec:expt:backbone}

To assess the generalizability of \algo{CoE}, we evaluate its effectiveness across three representative MLLM backbones: LLaVA-Next, InternVL2.5-8B, and Qwen2.5-VL-7B, on eight MMS benchmarks (Figure \ref{fig:backbone-result}).
%%%
Across all datasets and architectures, \algo{CoE} consistently outperforms all base models,  achieving improvements from \tb{+1.0} on VISTA to \tb{+5.1} on XMSMO.
These gains remain stable across varied content types, including news, instructional videos, sports commentary, and TV narratives.
%%%
The results confirm that \algo{CoE} functions as a \textbf{backbone-agnostic enhancement module}, providing consistent improvements in summarization regardless of model architecture.

\subsection{Effect of Model Size}
\label{sec:expt:modelsize}

To analyze scalability, we evaluate \algo{CoE} using Qwen2.5-VL backbones of different sizes (3B, 7B, 32B) and a proprietary GPT-5 model on four representative domains: \tb{Summ} (TV narratives), \tb{VIEWS} (news), \tb{BLiSS} (educational videos), and \tb{SoccerNet} (sports).
Each evaluation uses 100 randomly sampled test videos to ensure fairness and cost efficiency.

As shown in Figure~\ref{fig:model_size}, performance improves consistently with model scale: larger backbones (32B, GPT-5) deliver the best results, particularly on complex domains such as SoccerNet and BLiSS.
%%%
Minor deviations on Summ and VIEWS, where the 3B model slightly surpasses 7B, stem from its tendency toward extractive summarization with higher n-gram overlap, while larger models emphasize abstraction and coverage.
%%%
Overall, \algo{CoE} demonstrates strong scalability, maintaining competitive results with lightweight models and achieving state-of-the-art performance when paired with large-scale MLLMs.

\subsection{Case Study}
\label{sec:expt:case}

Figures \ref{fig:case_study_a} and \ref{fig:case_study_b} present qualitative examples on the \textbf{Summ} dataset, illustrating three core strengths of \algo{CoE}:
\begin{itemize}
  \item \textbf{Domain-adaptive Summarization (via DSG):} 
  In both cases (a) and (b), \algo{CoE} generates summaries that naturally reflect the domain's linguistic style, while \algo{CoS} tends to produce literal scene descriptions and fails to form coherent summaries, as seen in Case (b).

  \item \textbf{Precise Cross-modal Grounding (via HEG and CSG):}
  Guided by the HEG, \algo{CoE} accurately aligns sub-events with their corresponding video segments and grounds entities through fine-grained visual evidence, ensuring faithful cross-modal correspondence.

  \item \textbf{Robust Temporal Coherence (via EER):} 
  By modeling event evolution and entity transitions, \algo{CoE} constructs temporally coherent narratives that closely resemble human-written summaries.
\end{itemize}

These qualitative results demonstrate that \algo{CoE} moves beyond surface-level scene description toward event-centric, temporally coherent narratives that align closely with human-written summaries. 
%%%
Its strong robustness under domain shifts further demonstrates its suitability for real-world applications, including news reporting, sports analysis, and long-form episodic content.
%%%
Additional case studies across other datasets are provided in Appendix~\ref{sec:appen:case}, where consistent trends and observations can be observed.

%% file: tables/main_exprimental.tex
\definecolor{lightblue}{RGB}{173, 216, 230}
\definecolor{lightpink}{RGB}{255, 210, 230}
\definecolor{lightgray}{RGB}{211, 211, 211} 
\definecolor{lightgreen}{RGB}{197, 224, 180}
\definecolor{lightyellow}{RGB}{255, 255, 200} 
\definecolor{lightred}{RGB}{255, 182, 193}
\definecolor{lightpurple}{RGB}{216, 191, 216}

\definecolor{darkblue}{RGB}{123, 166, 180}  
\definecolor{darkpink}{RGB}{199, 21, 133}   
\definecolor{darkgray}{RGB}{169, 169, 169} 
\definecolor{darkgreen}{RGB}{94, 188, 94}    
\definecolor{darkyellow}{RGB}{218, 165, 32}   
\definecolor{darkred}{RGB}{205, 92, 92}    
\definecolor{darkpurple}{RGB}{166, 141, 166}

\begin{table}[t]
\centering
\small
\renewcommand{\arraystretch}{1.2}
\setlength{\tabcolsep}{3pt}
\resizebox{0.99\linewidth}{!}{
\begin{tabular}{ccccccccccc}
\toprule
    \multirow{2.5}{*}{\tb{Metric}} & \multirow{2.5}{*}{\tb{Method}} & \multicolumn{8}{c}{\tb{Dataset}} \\
    \cmidrule{3-10}
    & & \tb{VIEWS} & \tb{MM-AVS} & \tb{XMSMO} & \tb{TIB} & \tb{VISTA} & \tb{BLiSS} & \tb{SoccerNet} & \tb{Summ} \\
    \midrule

    % ---------------- BLEU-4 ----------------
    \multirow{5}{*}{\rotatebox{90}{\tb{BLEU-4 $\uparrow$}}} 
    & \algo{TCoT} & 1.78 & \ul{5.03} & \ul{1.67} & 1.17 & 2.65 & 0.44 & 0.00 & 4.27 \\
    & \algo{CoF} & \ul{4.09} & 4.98 & 0.51 & 1.66 & 3.69 & 0.42 & 0.42 & 1.04 \\
    & \algo{ViTCoT} & 3.50 & 4.26 & \ul{1.67} & 1.68 & \ul{4.28} & \ul{0.45} & \ul{1.46} & 0.80 \\
    & \algo{CoS} & 3.50 & 4.10 & 0.51 & \tb{2.24} & 4.03 & 0.31 & 1.27 & \tb{5.69} \\
    \cdashlinelr{2-10}
    & \cellcolor{lightpurple!25}\algo{CoE} & \cellcolor{lightpurple!25}\tb{5.36} & \cellcolor{lightpurple!25}\tb{6.02} & \cellcolor{lightpurple!25}\tb{4.08} & \cellcolor{lightpurple!25}\ul{2.13} & \cellcolor{lightpurple!25}\tb{4.80} & \cellcolor{lightpurple!25}\tb{0.53} & \cellcolor{lightpurple!25}\tb{15.44} & \cellcolor{lightpurple!25}\ul{4.59} \\

    \cmidrule{1-10}
    
    % ---------------- ROUGE ----------------
    \multirow{5}{*}{\rotatebox{90}{\tb{ROUGE $\uparrow$}}} 
    & \algo{TCoT} & 10.74 & 15.98 & \ul{15.32} & 11.37 & 13.85 & \ul{11.86} & 4.12 & 8.66 \\
    & \algo{CoF} & 16.86 & 14.17 & 8.64 & 12.36 & 14.20 & \tb{12.38} & 8.06 & 6.86 \\
    & \algo{ViTCoT} & 18.21 & 15.88 & 13.41 & 12.18 & 15.11 & 10.39 & \ul{13.94} & 9.10 \\
    & \algo{CoS} & \ul{18.73} & \ul{16.18} & 7.12 & \ul{13.29} & \ul{15.28} & 9.51 & 11.94 & \ul{11.88} \\
    \cdashlinelr{2-10}
    & \cellcolor{lightred!25}\algo{CoE} & \cellcolor{lightred!25}\tb{19.28} & \cellcolor{lightred!25}\tb{19.53} & \cellcolor{lightred!25}\tb{21.28} & \cellcolor{lightred!25}\tb{13.75} & \cellcolor{lightred!25}\tb{17.00} & \cellcolor{lightred!25}10.83 & \cellcolor{lightred!25}\tb{23.50} & \cellcolor{lightred!25}\tb{16.15} \\
    
    \cmidrule{1-10}
    \multirow{5}{*}{\rotatebox{90}{\tb{CIDEr $\uparrow$}}} 
    % ---------------- CIDEr ----------------
    & \algo{TCoT} & 3.81 & 7.14 & \ul{24.82} & 2.50 & 2.75 & \ul{1.87} & 0.00 & \ul{2.45} \\
    & \algo{CoF} & \ul{4.13} & \ul{7.49} & 1.60 & 2.50 & 2.12 & 1.24 & 0.03 & 0.09 \\
    & \algo{ViTCoT} & 1.56 & 4.32 & 10.61 & 1.85 & \ul{6.33} & 0.01 & \ul{1.48} & 0.19 \\
    & \algo{CoS} & 2.58 & 4.45 & 0.09 & \ul{3.27} & 3.01 & 0.05 & 0.57 & 1.27 \\
    \cdashlinelr{2-10}
    & \cellcolor{lightgreen!25}\algo{CoE} & \cellcolor{lightgreen!25}\tb{19.73} & \cellcolor{lightgreen!25}\tb{17.89} & \cellcolor{lightgreen!25}\tb{56.55} & \cellcolor{lightgreen!25}\tb{3.85} & \cellcolor{lightgreen!25}\tb{11.50} & \cellcolor{lightgreen!25}\tb{7.19} & \cellcolor{lightgreen!25}\tb{6.46} & \cellcolor{lightgreen!25}\tb{4.72} \\

    \cmidrule{1-10}
    % ---------------- METEOR ----------------
    \multirow{5}{*}{\rotatebox{90}{\tb{METEOR $\uparrow$}}} 
    & \algo{TCoT} & 5.99 & 10.17 & 12.21 & 6.69 & 10.63 & 8.50 & 1.12 & 5.73 \\
    & \algo{CoF} & 12.98 & 9.89 & 8.80 & 2.89 & \ul{13.91} & \ul{8.92} & 3.64 & 3.95 \\
    & \algo{ViTCoT} & \ul{16.26} & 13.10 & \ul{14.47} & \tb{12.22} & 13.71 & \tb{9.91} & \ul{9.41} & 7.37 \\
    & \algo{CoS} & \tb{16.57} & \ul{14.26} & 8.29 & 9.86 & 13.16 & 8.38 & 6.13 & \ul{8.38} \\
    \cdashlinelr{2-10}
    & \cellcolor{lightyellow!25}\algo{CoE} & \cellcolor{lightyellow!25}12.63 & \cellcolor{lightyellow!25}\tb{14.51} & \cellcolor{lightyellow!25}\tb{17.24} & \cellcolor{lightyellow!25}\ul{10.34} & \cellcolor{lightyellow!25}\tb{16.75} & \cellcolor{lightyellow!25}6.81 & \cellcolor{lightyellow!25}\tb{21.35} & \cellcolor{lightyellow!25}\tb{12.29} \\

    \cmidrule{1-10}
    % ---------------- BERTScore ----------------
    \multirow{5}{*}{\rotatebox{90}{\tb{BERTScore $\uparrow$}}} 
    & \algo{TCoT} & 85.68 & \ul{85.69} & \ul{86.80} & \ul{83.85} & \ul{84.38} & 83.23 & 80.88 & \ul{82.33} \\
    & \algo{CoF} & 85.83 & 82.99 & 80.77 & 83.72 & 83.03 & \tb{84.93} & 80.50 & 79.33 \\
    & \algo{ViTCoT} & 86.61 & 84.28 & 86.50 & 82.62 & 83.78 & 82.30 & 81.14 & 79.56 \\
    & \algo{CoS} & \ul{86.65} & 83.93 & 83.91 & 83.65 & 83.78 & \ul{83.97} & \ul{81.22} & 81.86 \\
    \cdashlinelr{2-10}
    & \cellcolor{lightblue!25}\algo{CoE} & \cellcolor{lightblue!25}\tb{88.26} & \cellcolor{lightblue!25}\tb{88.62} & \cellcolor{lightblue!25}\tb{88.65} & \cellcolor{lightblue!25}\tb{84.91} & \cellcolor{lightblue!25}\tb{85.87} & \cellcolor{lightblue!25}83.80 & \cellcolor{lightblue!25}\tb{84.99} & \cellcolor{lightblue!25}\tb{85.76} \\

    \cmidrule{1-10}
    % ---------------- F1-score ----------------
    \multirow{5}{*}{\rotatebox{90}{\tb{F1-score $\uparrow$}}} 
    & \algo{TCoT} & 18.42 & 24.47 & \ul{28.44} & 14.27 & 17.18 & 2.37 & 0.94 & 15.00 \\
    & \algo{CoF} & 29.64 & 19.49 & 27.67 & 18.51 & 17.62 & 3.07 & 19.52 & 2.46 \\
    & \algo{ViTCoT} & \ul{34.93} & \ul{26.80} & 27.86 & \ul{22.12} & \ul{20.76} & \tb{4.44} & 25.77 & 2.51 \\
    & \algo{CoS} & 32.59 & 26.25 & 27.16 & 21.87 & 16.97 & 2.75 & \ul{31.45} & \ul{31.06} \\
    \cdashlinelr{2-10}
    & \cellcolor{lightgray!25}\algo{CoE} & \cellcolor{lightgray!25}\tb{38.14} & \cellcolor{lightgray!25}\tb{33.91} & \cellcolor{lightgray!25}\tb{33.82} & \cellcolor{lightgray!25}\tb{25.34} & \cellcolor{lightgray!25}\tb{29.01} & \cellcolor{lightgray!25}\ul{3.63} & \cellcolor{lightgray!25}\tb{63.37} & \cellcolor{lightgray!25}\tb{69.50} \\

    \cmidrule{1-10}
    \multirow{5}{*}{\rotatebox{90}{\tb{G-Eval $\uparrow$}}} 
    % ---------------- G-Eval-score ----------------
    & \algo{TCoT}   & 2.27      & \ul{3.31} & 3.61      & 2.79      & 3.12      & 2.57      & 1.95      & 1.95 \\
    & \algo{CoF}    & 2.57      & 2.47      & 3.03      & 2.91      & 2.79      & 2.70      & 2.21      & 1.63 \\
    & \algo{ViTCoT} & \ul{3.34} & 3.24      & \tb{3.81} & \ul{3.32} & \ul{3.42} & \ul{2.94} & \tb{2.75} & 2.00 \\
    & \algo{CoS}    & 3.04      & 2.89      & 2.99      & 2.85      & 2.68      & 2.41      & 2.52      & \ul{2.13} \\
    \cdashlinelr{2-10}
    & \cellcolor{lightpink!25}\algo{CoE}    & \cellcolor{lightpink!25}\tb{3.44} & \cellcolor{lightpink!25}\tb{4.01} & \cellcolor{lightpink!25}\ul{3.65} & \cellcolor{lightpink!25}\tb{3.58} & \cellcolor{lightpink!25}\tb{3.82} & \cellcolor{lightpink!25}\tb{3.00} & \cellcolor{lightpink!25}\ul{2.70} & \cellcolor{lightpink!25}\tb{3.16} \\
    \bottomrule
\end{tabular}}
\vspace{-0.5em}
\caption{\tb{Comparison with Video CoT Baselines.} Quantitative results of \algo{CoE} and competing video CoT methods on eight MMS datasets, all evaluated in a zero-shot, training-free setting without dataset-specific fine-tuning. For each dataset-metric pair, \tb{bold} and \ul{underline} denote the best and second-best results, respectively.}
% \caption{Comparison of \algo{CoE} with video CoT baselines on eight MMS datasets. All methods are evaluated in a training-free, zero-shot setting without any dataset-specific fine-tuning. For each dataset–metric pair, \tb{bold} and \ul{underline} denote the best and second-best results, respectively.}
\label{tab:main-results}
\end{table}
\setlength{\textfloatsep}{1.0em}

%% file: tables/ablation_rouge.tex
\definecolor{lightblue}{RGB}{173, 216, 230}
\definecolor{lightpink}{RGB}{255, 182, 193}  % 修改为更接近粉色
\definecolor{lightgray}{RGB}{211, 211, 211}  % 修改为更接近灰色
\definecolor{lightgreen}{RGB}{197, 224, 180}
\definecolor{lightyellow}{RGB}{255, 255, 200}  % 调整为更亮的黄色
\definecolor{lightred}{RGB}{255, 182, 193}
\definecolor{lightpurple}{RGB}{216, 191, 216}

\definecolor{darkblue}{RGB}{123, 166, 180}  
\definecolor{darkpink}{RGB}{199, 21, 133}   % 修改为深粉色
\definecolor{darkgray}{RGB}{169, 169, 169}  % 修改为深灰色
\definecolor{darkgreen}{RGB}{94, 188, 94}    
\definecolor{darkyellow}{RGB}{218, 165, 32}   % 调整为更深的黄色
\definecolor{darkred}{RGB}{205, 92, 92}     % 这是稍微深一点的红色
\definecolor{darkpurple}{RGB}{166, 141, 166}

\begin{table}[t]
\centering
\small
\vspace{-0.5em}
\renewcommand{\arraystretch}{1.25}
\setlength{\tabcolsep}{3pt}  % 调整列间距
\resizebox{0.99\linewidth}{!}{
\begin{tabular}{*{9}{c}}
    \toprule
    \multirow{2.5}{*}{\tb{Method}} & \multicolumn{8}{c}{\tb{Dataset}} \\
    \cmidrule{2-9}
    & \textbf{VIEWS} & \textbf{MM-AVS} & \textbf{XMSMO} & \textbf{TIB} & \textbf{VISTA} & \textbf{BLiSS} & \textbf{SoccerNet} & \textbf{Summ} \\
    
    \midrule
    \tb{CoE -- HEG} & 18.68      & 16.68      & 19.82      & 12.80       & \ul{16.77} & 10.44       & 22.48      & \tb{16.25} \\
    \tb{CoE -- CSG} & 18.56      & 16.35      & 20.79      & 13.16       & 16.56      & 9.91        & \ul{23.16} & \ul{16.20} \\
    \tb{CoE -- EER} & 18.66      & \ul{19.44} & 20.88      & 13.04       & 16.53      & 9.99        & 22.44      & 15.90      \\
    \tb{CoE -- DSG} & \ul{18.76} & 18.70      & \ul{21.17} & \ul{13.38}  & 16.75      & \ul{10.80}  & 22.55      & 15.67      \\
    \cdashlinelr{1-9}
    \cellcolor{lightred!25}\algo{CoE} & \cellcolor{lightred!25}\tb{19.28} & \cellcolor{lightred!25}\tb{19.53} & \cellcolor{lightred!25}\tb{21.28} & \cellcolor{lightred!25}\tb{13.75} & \cellcolor{lightred!25}\tb{17.00} & \cellcolor{lightred!25}\tb{10.83} & \cellcolor{lightred!25}\tb{23.50} & \cellcolor{lightred!25}16.15 \\
    \bottomrule
\end{tabular}}
% \vspace{-0.5em}
\caption{\tb{Ablation Study.} ROUGE scores of \algo{CoE} and its ablated variants across eight benchmark datasets.}
\label{tab:ablation-rouge}
\end{table}

%% file: tables/ablation_cider.tex
\definecolor{lightblue}{RGB}{173, 216, 230}
\definecolor{lightpink}{RGB}{255, 182, 193}  % 修改为更接近粉色
\definecolor{lightgray}{RGB}{211, 211, 211}  % 修改为更接近灰色
\definecolor{lightgreen}{RGB}{197, 224, 180}
\definecolor{lightyellow}{RGB}{255, 255, 200}  % 调整为更亮的黄色
\definecolor{lightred}{RGB}{255, 182, 193}
\definecolor{lightpurple}{RGB}{216, 191, 216}

\definecolor{darkblue}{RGB}{123, 166, 180}  
\definecolor{darkpink}{RGB}{199, 21, 133}   % 修改为深粉色
\definecolor{darkgray}{RGB}{169, 169, 169}  % 修改为深灰色
\definecolor{darkgreen}{RGB}{94, 188, 94}    
\definecolor{darkyellow}{RGB}{218, 165, 32}   % 调整为更深的黄色
\definecolor{darkred}{RGB}{205, 92, 92}     % 这是稍微深一点的红色
\definecolor{darkpurple}{RGB}{166, 141, 166}

\begin{table}[t]
\centering
\small
\renewcommand{\arraystretch}{1.25}
\setlength{\tabcolsep}{3pt}  % 调整列间距
\resizebox{0.99\linewidth}{!}{
\begin{tabular}{*{9}{c}}
    \toprule
    \multirow{2.5}{*}{\tb{Method}} & \multicolumn{8}{c}{\tb{Dataset}} \\
    \cmidrule{2-9}
    & \textbf{VIEWS} & \textbf{MM-AVS} & \textbf{XMSMO} & \textbf{TIB} & \textbf{VISTA} & \textbf{BLiSS} & \textbf{SoccerNet} & \textbf{Summ} \\
    \midrule
    \tb{CoE -- HEG} & 17.20      & \ul{17.01} & \ul{44.09} & 1.33      & \ul{10.16} & \tb{8.02}  & 5.46      & 1.89      \\
    \tb{CoE -- CSG} & 17.78      & 16.27      & 40.47      & 1.44      & 9.98       & 7.03       & 4.60      & 3.91      \\
    \tb{CoE -- EER} & \ul{18.54} & 13.83      & 43.58      & \ul{1.60} & 9.37       & \ul{7.21}  & 4.23      & \ul{4.54} \\
    \tb{CoE -- DSG} & 9.23       & 12.85      & 40.56      & 1.51      & 9.55       & 3.70       & \ul{6.00} & 2.14      \\
    \cdashlinelr{1-9}
    \cellcolor{lightgreen!25}\tb{CoE} & \cellcolor{lightgreen!25}\tb{19.73} & \cellcolor{lightgreen!25}\tb{17.89} & \cellcolor{lightgreen!25}\tb{56.55} & \cellcolor{lightgreen!25}\tb{3.85} & \cellcolor{lightgreen!25}\tb{11.50} & \cellcolor{lightgreen!25}7.19 & \cellcolor{lightgreen!25}\tb{6.46} & \cellcolor{lightgreen!25}\tb{4.72} \\
    \bottomrule
\end{tabular}
}
% \vspace{-0.5em}
\caption{\tb{Ablation Study.} CIDEr scores of \algo{CoE} and its ablated variants across eight benchmark datasets.}
\label{tab:ablation-cider}
\end{table}

% \begin{table*}[t]
% \centering
% \small
% \renewcommand{\arraystretch}{1.1}
% % \setlength{\tabcolsep}{8pt}  % 调整列间距
% % \captionsetup{skip=0.5em}
% % \resizebox{0.99\textwidth}{!}{
% \begin{tabular}{*{4}c|*{8}{c}}
%     \toprule
%     \textbf{HEG} & \textbf{CSG} & \textbf{EER} & \textbf{ST} & \textbf{VIEWS} & \textbf{MM-AVS} & \textbf{XMSMO} & \textbf{TIB} & \textbf{VISTA} & \textbf{BLiSS} & \textbf{SoccerNet} & \textbf{Summ} \\
%     \midrule
%     \XSolidBrush & \Checkmark   & \Checkmark    & \Checkmark & 17.20 & 17.01 & 44.09 & 1.33 & 10.16 & 8.02 & 5.46 & 1.89 \\
%     \Checkmark   & \XSolidBrush & \Checkmark    & \Checkmark    & 17.78 & 16.27 & 40.47 & 1.44 & 9.98 & 7.03 & 4.60 & 3.91 \\
%     \Checkmark   & \Checkmark   & \XSolidBrush  & \Checkmark    & 18.54 & 13.83 & 43.58 & 1.60 & 9.37 & 7.21 & 4.23 & 4.54 \\
%     \Checkmark   & \Checkmark   & \Checkmark    & \XSolidBrush  & 9.23 & 12.85 & 40.56 & 1.51 & 9.55 & 3.70 & 6.00 & 2.14 \\
%     \midrule
%     \Checkmark   & \Checkmark   & \Checkmark    & \Checkmark    & 19.73 & 17.89 & 56.55 & 3.85 & 11.50 & 7.19 & 6.46 & 4.72 \\
    
%     \bottomrule
% \end{tabular}
% % }
% \caption{CIDEr scores of our method on various vision–language backbones across eight benchmark datasets. Models marked with $^{\star}$ are enhanced with the proposed HERMSA framework. The results demonstrate the strong generalization ability and robustness of our approach across diverse backbone architectures.}
% \label{tab:ablation-results}
% \end{table*}

%% file: 06_conclusions.tex
\section{Conclusions and Future Work}
\label{sec:conclusions}

In this work, we introduce \algo{CoE}, a training-free framework for multimodal summarization that replaces implicit feature fusion with explicit event-centric reasoning.
%%%
\algo{CoE} organizes video-text inputs into a hierarchical event representation and performs grounded spatio-temporal reasoning to generate coherent, entity-aware summaries, while a lightweight style adaptation module aligns the results with domain-specific linguistic conventions.
%%%
Extensive experiments on eight diverse MMS benchmarks show that \algo{CoE} consistently outperforms strong video CoT baselines in zero-shot settings, achieving substantial improvements across both lexical and semantic metrics.
%%%
These results demonstrate that explicit event-centric reasoning effectively mitigates the challenges of supervision-heavy training, weak cross-modal grounding, and flat temporal modeling, offering a principled, scalable, and practical approach to real-world multimodal summarization.

Looking forward, \algo{CoE}'s structured reasoning paradigm can be naturally extended to multimodal output generation (e.g., key frame selection, video highlights) and interactive summarization, where users specify events of interest, paving the way for interpretable and adaptable multimodal summarization systems in real-world applications.

\section*{Acknowledgements}
We sincerely thank the anonymous reviewers for their insightful and constructive feedback, which greatly improved the quality of this paper.
This work was supported by the National Natural Science Foundation of China (NSFC) under Grant Nos. 62125201, U24B20174, and U25B6003.

%% file: 07_appendix.tex
\clearpage
\setcounter{page}{1}
\setcounter{figure}{0}
\setcounter{table}{0}
\renewcommand{\thefigure}{A\arabic{figure}}
\renewcommand{\thetable}{A\arabic{table}}
\maketitlesupplementary
\appendix

% \section{Appendix}
\label{sec:app}

\section{Generalization Test}
\label{sec:appen:general}

To systematically assess the generalization capability of Chain-of-Events (\algo{CoE}), we conduct extensive cross-domain experiments across other seven diverse MMS benchmarks: BLiSS~\cite{he2023bliss} (instructional videos), MM-AVS~\cite{fu2021mm-avs} and XMSMO~\cite{tang2023tldw} (news reports), SoccerNet~\cite{mkhallati2023soccernet} (sports broadcasts), Summ~\cite{papalampidi2023hierarchical3d} (TV series), and TIB~\cite{gigant2023tib} and VISTA~\cite{liu2025vista} (lecture videos). These datasets exhibit substantial domain variations in content structure, visual semantics, and temporal dynamics.

\begin{figure*}[t]
  \centering
  \includegraphics[width=0.99\textwidth]{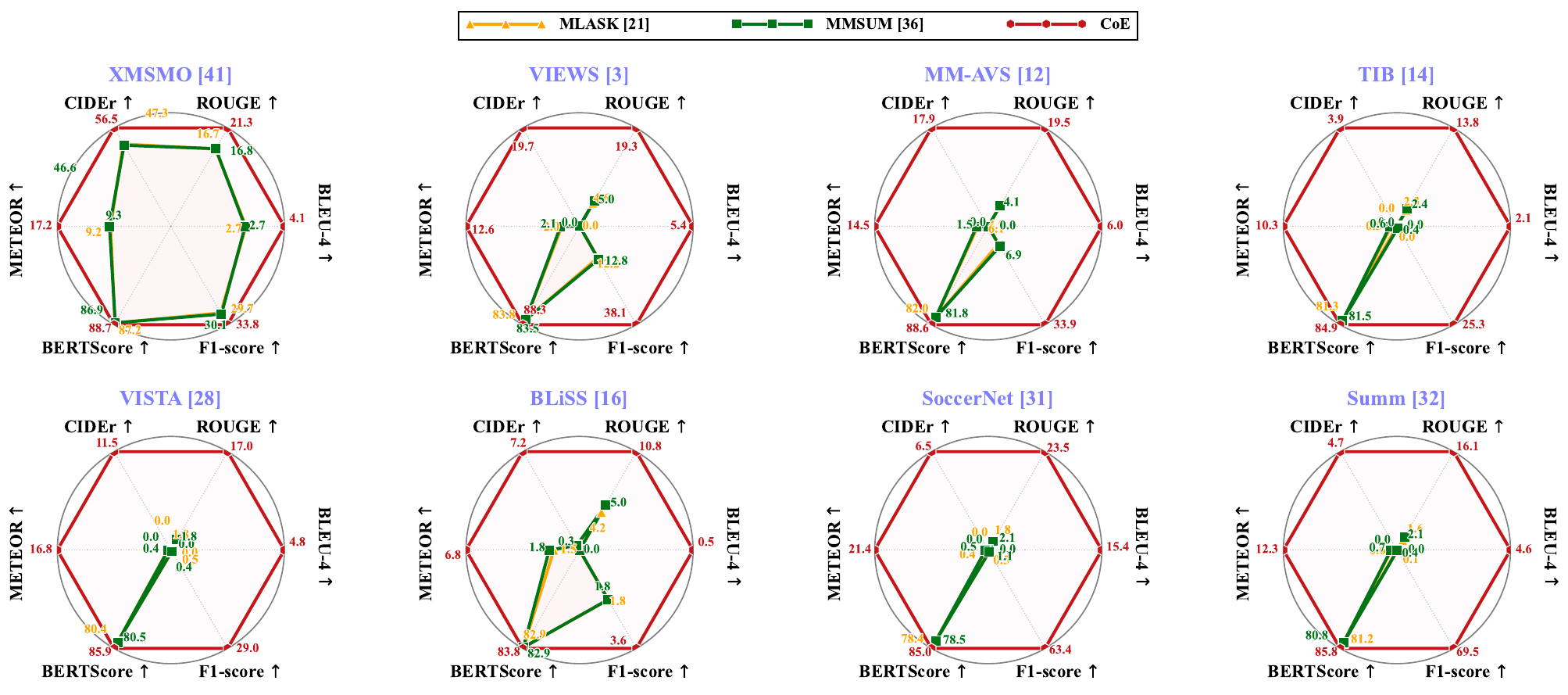}
  \vspace{-0.5em}
  \caption{\textbf{Motivating Experiments on XMSMO.} Existing MMS models (e.g., \algo{MLASK}~\cite{krubinski2023mlask} and \algo{MMSum}~\cite{qiu2024mmsum}) achieve strong in-domain results when trained on XMSMO~\cite{tang2023tldw}, but their performance drops sharply under domain shift. In contrast, our \textbf{training-free} \algo{CoE} framework generalizes effectively across diverse datasets, maintaining stable zero-shot performance without task-specific training or adaptation.}
  \label{fig:motivation-xmsmo}
  \vspace{-0.5em}
\end{figure*}

\begin{figure*}[t]
  \centering
  \includegraphics[width=0.99\textwidth]{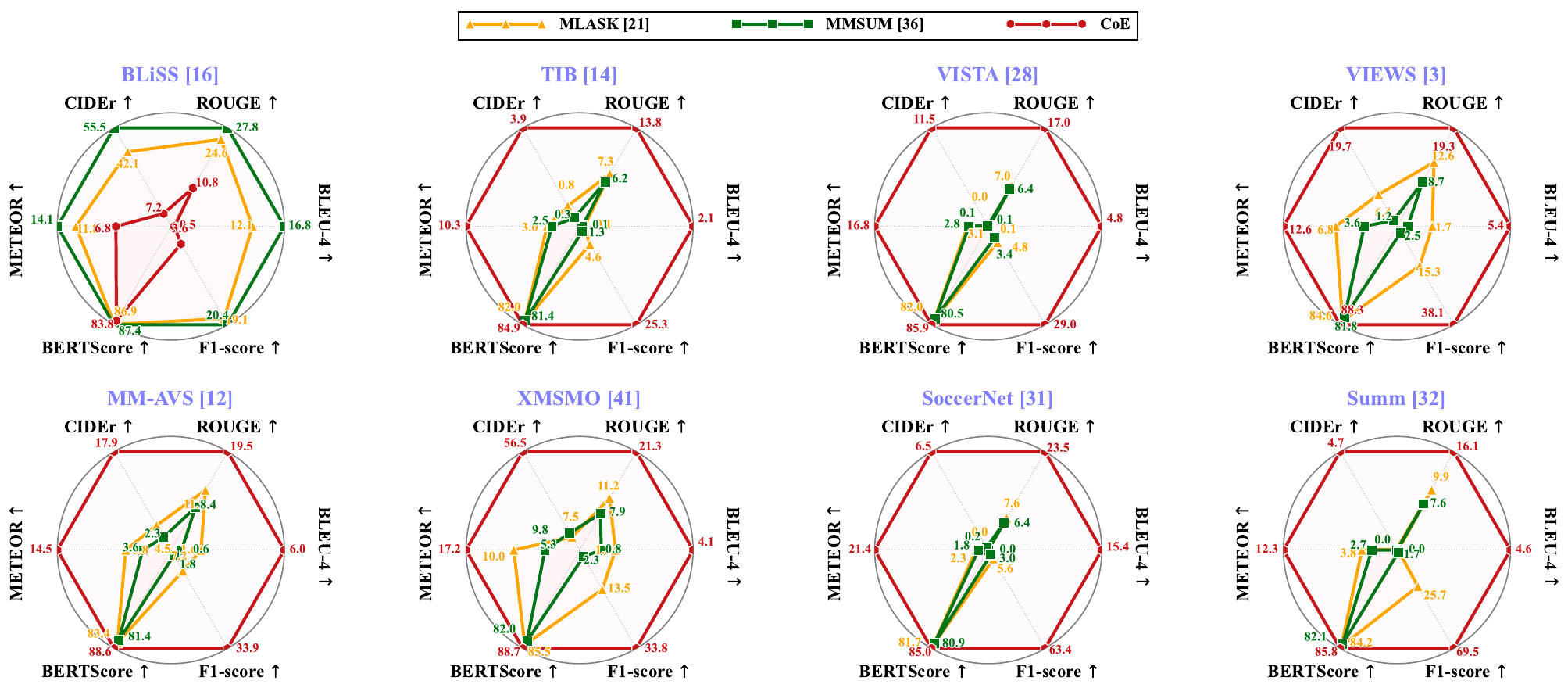}
  \vspace{-0.5em}
  \caption{\textbf{Motivating Experiments on BLiSS.} Existing MMS models (e.g., \algo{MLASK}~\cite{krubinski2023mlask} and \algo{MMSum}~\cite{qiu2024mmsum}) achieve strong in-domain results when trained on BLiSS~\cite{he2023bliss}, but their performance drops sharply under domain shift. In contrast, our \textbf{training-free} \algo{CoE} framework generalizes effectively across diverse datasets, maintaining stable zero-shot performance without task-specific training or adaptation.}
  \label{fig:motivation-bliss}
  \vspace{-0.5em}
\end{figure*}

For each dataset, we train \algo{MLASK}\cite{krubinski2023mlask} and \algo{MMSum}\cite{qiu2024mmsum} baselines using their official implementations, then evaluate them on all remaining datasets without fine-tuning or domain adaptation. 
%%%
In contrast, \algo{CoE} operates in a purely zero-shot manner across all benchmarks, requiring no task-specific training. 
%%%
As shown in Figures~\ref{fig:motivation-bliss}--\ref{fig:motivation-xmsmo}, the results exhibit highly consistent trends:
\begin{itemize}[nolistsep]
    \item \tb{In-domain overfitting of supervised methods.} Models trained on individual datasets perform well within their training domain but experience substantial performance drops when evaluated on different domains. For instance, a model trained on BLiSS shows strong results on instructional videos but struggles with sports footage from SoccerNet or scripted dialogues from SummScreen. This degradation occurs regardless of which dataset is used for training, revealing a fundamental limitation: existing fusion-based methods tend to overfit to domain-specific characteristics such as visual style, narrative pacing, or vocabulary distribution.

    \item \tb{Consistent zero-shot transfer with CoE.} Unlike supervised baselines, our training-free framework maintains steady performance across all datasets, while baseline methods exhibit unstable patterns when tested out-of-domain. This stability arises naturally from \algo{CoE}'s design, which adapts to different video genres through hierarchical event modeling rather than relying on dataset-specific patterns learned during training.

    \item \tb{Competitive performance even without in-domain supervision.}
    \algo{CoE} also achieves \emph{comparable or even better} scores than supervised baselines on several datasets (e.g., MM-AVS, BLiSS, Summ). These observations indicate that \tb{CoE's} event-centric architecture, which combines hierarchical event graph construction, cross-modal grounding, event-evolution reasoning, and lightweight style adaptation, provides a strong inductive bias that alleviates the need for task-specific supervision while still yielding competitive performance on target domains.

\end{itemize}

In summary, these cross-domain experiments show that existing supervised MMS models are strongly tied to their training domains, whereas \algo{CoE} delivers stable zero-shot performance across heterogeneous benchmarks. By relying on an event-centric architecture instead of dataset-specific supervision, \algo{CoE} offers a more domain-agnostic solution for MMS and remains competitive even on datasets where other methods are explicitly trained.

\section{Prompts}
\label{sec:appen:prompt}

\subsection{Hierarchical Event Graph~(HEG) Construction}
\algo{CoE} first decomposes the input text to construct a hierarchical event graph, which contains three layers: a global event layer, a sub-event layer, and an entity-relation layer.
%%%
The global event captures the overall theme of the narrative, while sub-events decompose the global event into semantically coherent components. 
%%%
We use the following prompt to jointly infer the main event and its sub-events:

\begin{promptbox}
    Analyze this video transcript or article and provide: \\

    \hspace*{1em}1. A main event summary in one sentence that captures the high-level essence of the entire video. Keep it concise, preferably under 50 words. \\

    \hspace*{1em}2. Determine if the video can be meaningfully divided into sub-events (maximum 3). If yes, provide concise one-sentence summaries for each sub-event. Keep the granularity appropriate - don't make the sub-events too specific or detailed. \\

    Return your response in JSON format: \\
    If the transcript CANNOT be divided into sub-events: \\
    \hspace*{1em}\{``main\_event": ``your main summary here", ``sub\_events": [main event]\} \\
    If the transcript CAN be divided into sub-events: \\
    \hspace*{1em}\{``main\_event": ``your main summary here", ``sub\_events": [``sub-event 1", ``sub-event 2", ...]\} \\

    Here are two examples of good main event summaries: \\
    \hspace*{1em}- \{Example 1\} \\
    \hspace*{1em}- \{Example 2\} \\

    The following is the video transcript: \{input text\}
\end{promptbox}

% Then, for each sub-event, \algo{CoE} extracts the entity set and relations to form a sub graph.
Then, for each sub-event, \algo{CoE} refines its representation at the entity-relation layer by extracting an event-specific set of typed entities and organizing them into a subgraph. Concretely, we prompt the MLLM to identify all entities that are directly relevant to the given sub-event, including (i) \emph{persons} involved in or mentioned within the event, (ii) \emph{locations} where the event takes place, (iii) \emph{organizations} such as companies, institutions, or teams, and (iv) salient \emph{objects/items} that play a role in the event. All entities are required to be explicitly grounded in the transcript and deduplicated. These event-specific entities serve as the nodes of the entity-relation subgraph, which is later used to provide fine-grained semantic anchors for cross-modal grounding and reasoning.

\begin{promptbox}
    Please extract relevant entities related to the specified event from the provided video transcript. Entities should include people, locations, organizations, and objects/items. Return the results in JSON format with the following structure: \\
    \{ \\
        \hspace*{1em}"person": ["name 1", "name 2", "name 3"], \\
        \hspace*{1em}"location": ["place 1", "place 2", "place 3"], \\
        \hspace*{1em}"organization": ["org 1", "org 2", "org 3"], \\
        \hspace*{1em}"item": ["item 1", "item 2", "item 3"] \\
    \} \\
    
    Instructions: \\
    \hspace*{1em}1. Only include entities directly related to the specified event \\
    \hspace*{1em}2. Remove duplicates and normalize entity names \\
    \hspace*{1em}3. For people, use full names when available \\
    \hspace*{1em}4. For organizations, use official names rather than abbreviations when possible \\
    \hspace*{1em}5. Ensure all extracted entities are actually mentioned in the transcript \\
    \hspace*{1em}6. If no entities of a certain type are found, return an empty array for that category \\

    Event: \{event\} \\
    Video transcript: \{text\} \\
\end{promptbox}

In summary, the HEG construction module transforms unstructured textual input into a structured semantic scaffold by sequentially decomposing the narrative into global events, sub-events, and fine-grained entity sets. By organizing information across these three hierarchical layers, we capture both the high-level thematic evolution and the precise entity-level details necessary for reasoning. This constructed graph serves as a robust domain-agnostic prior, providing the essential semantic anchors that guide the subsequent cross-modal grounding and event evolution analysis.

\subsection{Cross-modal Spatial Grounding (CSG)}
In the CSG module, we first perform a sub-event alignment procedure grounded in the HEG. Given an input video, we uniformly sample a fixed number of frames and partition them into shot-level clips, each of which encapsulates a coherent local temporal context. Guided by the HEG, every clip is then associated with its most relevant sub-event. The prompt employed in this process is defined as follows:
\begin{promptbox}[slightgreen]
    Please analyze the \{domain\} scene in this set of images and determine which of the following sub-events this video clip belongs to. \\
    
    Return the result in JSON format with three fields: \\
    \hspace*{1em}- ``idx": the chosen sub-event number (starting from 0), \\
    \hspace*{1em}- ``sub-event": the description of the chosen sub-event, \\
    \hspace*{1em}- ``reasoning": a short explanation of why this sub-event was selected based on the video content. \\
    
    The JSON format must be:  \{  ``idx": number,  ``sub-event": ``event description",  ``reasoning": ``short explanation"\} \\
    
    Sub-events: \{sub-event list\}
\end{promptbox}

Following the assignment of clips to specific sub-events, we proceed to fine-grained spatial grounding. 
%%%
In this step, the goal is to construct a visually grounded subgraph by verifying which entities are visible in the current clip and identifying their interactions. 
%%%
The prompt designed to extract these visual entity-relation triples is provided below:
\begin{promptbox}[slightgreen]
    Extract entity relationships from the video by following these steps: \\
    
    \hspace*{1em}1. List visible entities: What people, organizations, or locations appear?\\
    \hspace*{1em}2. Identify actions/relationships: What connections exist between entities?\\
    \hspace*{1em}3. Match with subgraph: Find corresponding triples in the provided subgraph\\
    \hspace*{1em}4. Verify relevance: Ensure matches are reasonable (direct or contextual)\\
    
    If no exact match exists, select the most contextually relevant triple.\\
    
    Subgraph: \{subgraph\} \\
    
    Output format: \{``reasoning": ``your analysis", ``triples": [\{``from": ``David Warner", ``relation": ``celebrated his century at", ``to": ``Coogee Oval"\}]\}
\end{promptbox}

Through this two-stage prompting strategy, the CSG module effectively transforms the raw video stream into a sequence of visually grounded subgraphs. 
%%%
By enforcing both temporal alignment with sub-events and spatial verification of entity relations, we ensure that the reasoning process is anchored in concrete visual evidence rather than relying solely on textual priors. 
%%%
This yields fine-grained, reliable correspondences between visual observations and textual concepts, providing a robust structured representation for the subsequent EER module.

\subsection{Event Evolution Reasoning (EER)}
% Building on the Video Clip Aggregation step~\ref{sec:method:event-evolution}, the EER module focuses on capturing the dynamic evolution of the narrative. 
% %%%
% Instead of treating segments in isolation, this module analyzes the causal and temporal dependencies between them. 

% Specifically, for each aggregated temporal segment, the model is provided with the global textual context, the segment's visual content, and a structural comparison between the current entity-relation graph and that of the preceding segment. 
% %%%
% This setup allows the model to explicitly track the trajectory of the event and to identify how entities persist, interact, or change over time, instead of merely describing static scenes.

Building on the Video Clip Aggregation step~\ref{sec:method:event-evolution}, the EER module focuses on capturing the dynamic evolution of the narrative. 
%%%
Instead of treating segments in isolation, this module analyzes the causal and temporal dependencies between them. 

Specifically, for each aggregated temporal segment, the model is provided with the global textual context, the segment's visual content, and a structural comparison between the current entity-relation graph and that of the preceding segment. 
%%%
By contrasting the added and removed entities or relations across adjacent segments, the model can distinguish between mere visual redundancy and genuine narrative transitions . 
%%%
This setup allows the model to explicitly track the trajectory of the event and to identify how entities persist, interact, or change over time, instead of merely describing static scenes, which in turn provides a more stable event backbone for the subsequent summary generation module.

The prompt used to reason about these transitions and generate the trajectory description is as follows:
\begin{promptbox}[slightyellow]
    Based on the reference transcript, the given event and sub-event graph, and the newly identified entities and relations, extract the most relevant information from the transcript. Then analyze the event trajectory and generate a concise description in no more than 100 words.  \\
    
    STRICT OUTPUT FORMAT (no extra text):  \\
    
    \hspace*{1em}Event trajectory: \{analysis of event progression\} \\
    \hspace*{1em}Description summary: \{concise summary\}\\
    
    Inputs:  
    
    \hspace*{1em}- Reference transcript: {article}\\
    \hspace*{1em}- Sub-event: {sub-event}\\
    \hspace*{1em}- Subgraph: {subgraph}\\
    \hspace*{1em}- New entities and relations: \{New entities and relations\}
\end{promptbox}

\subsection{Domain-adaptive Summary Generation (DSG)}
The final module, DSG, operates in two distinct phases to synthesize the final output. 
%%%
First, we construct a comprehensive initial summary by aggregating the sequence of event trajectory descriptions produced by the EER module. 
%%%
Rather than simply concatenating these descriptions, we prompt the MLLM to consolidate the key narrative developments, causal transitions, and entity interactions into a unified and coherent text. 

This initial draft serves as a content-heavy backbone, ensuring that all salient multimodal information extracted during the reasoning process is preserved. 
%%%
The prompt used to generate this event-centric initial summary is as follows:
\begin{promptbox}[slightred]
    Generate a \{domain\} summary following these steps:\\
    1. Identify key information: \\
    \hspace*{1em}- Main event and participants from the article \\
    \hspace*{1em}- Timeline and locations from scene descriptions 
    \hspace*{1em}- Critical facts and outcomes \\
    2. Structure the summary: \\
    \hspace*{1em}- Lead with the most important fact \\
    \hspace*{1em}- Follow with supporting details \\
    \hspace*{1em}- Keep similar length to examples \\
    3. Match example style: \\
    \hspace*{1em}- Concise, fact-focused sentences\\
    \hspace*{1em}- No commentary or interpretation \\
    \hspace*{1em}- Include names, numbers, specific details\\ 
    
    Examples: 
    
    \hspace*{1em}\{Example 1\}\\
    \hspace*{1em}\{Example 2\}\\
    \hspace*{1em}\{Example 3\}\\
    
    Overall event: \{total event\}\\
    Sub-events: \{sub-events\}\\
    Entities and relations: \{entities and relations\}\\
    Scene descriptions: \{scene descriptions\}\\
    News Article: \{article\}
\end{promptbox}

While the initial summary captures the factual content, it may lack the specific linguistic nuances of the target domain. 
%%%
To address this, the second phase employs a lightweight style adaptation mechanism. 
%%%
We retrieve a small set of reference summaries from the target domain to serve as stylistic exemplars. 
%%%
The MLLM is then prompted to rewrite the initial summary, aligning its tone, phrasing, and discourse structure with these exemplars while strictly maintaining content fidelity. 
%%%
The prompt for this style refinement process is provided below:

\begin{promptbox}[slightred]
    Rewrite the text by thinking through these steps: \\
    1. Analyze examples' characteristics: \\
    - Sentence structure (short, fact-dense)\\ 
    - Information density (3-5 key facts per example) \\
    - Length range (estimate word count from examples) \\
    
    2. Extract key facts from input text: \\
    - Who, what, when, where, why \\
    - Remove redundancy and commentary \\
    
    3. Reconstruct in example style \\
    - Use compact sentences \\
    - Maintain factual accuracy \\
    - Match length to example \\
    
    Examples (style and length reference): \\
    \{Example 1\}\\
    \{Example 2\}\\
    \{Example 3\}\\
    
    Text: \{initial summary\}
\end{promptbox}

By decoupling content synthesis from stylistic refinement, the DSG module ensures robust generalization across diverse benchmarks.
%%%
This separation allows \algo{CoE} to maintain high factual accuracy while flexibly adapting its linguistic output to match the distinct conventions of news, sports, or instructional videos without requiring domain-specific fine-tuning.

\begin{figure*}[t]
  \centering
  \includegraphics[width=0.99\textwidth]{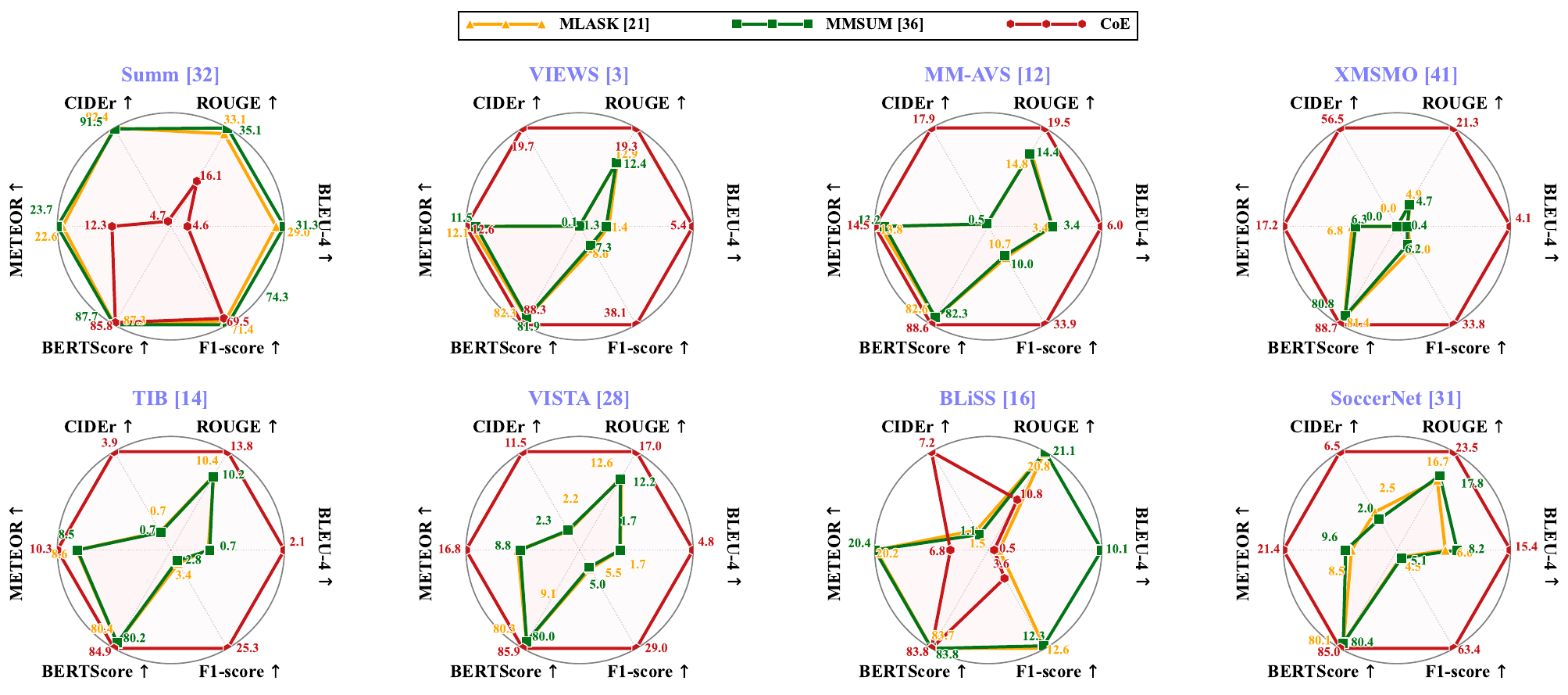}
  \vspace{-0.5em}
  \caption{\textbf{Motivating Experiments on Summ.} Existing MMS models (e.g., \algo{MLASK}~\cite{krubinski2023mlask} and \algo{MMSum}~\cite{qiu2024mmsum}) achieve strong in-domain results when trained on Summ~\cite{papalampidi2023hierarchical3d}, but their performance drops sharply under domain shift. In contrast, our \textbf{training-free} \algo{CoE} framework generalizes effectively across diverse datasets, maintaining stable zero-shot performance without task-specific training or adaptation.}
  \label{fig:motivation-summ}
  \vspace{-0.5em}
\end{figure*}

\begin{figure*}[t]
  \centering
  \includegraphics[width=0.99\textwidth]{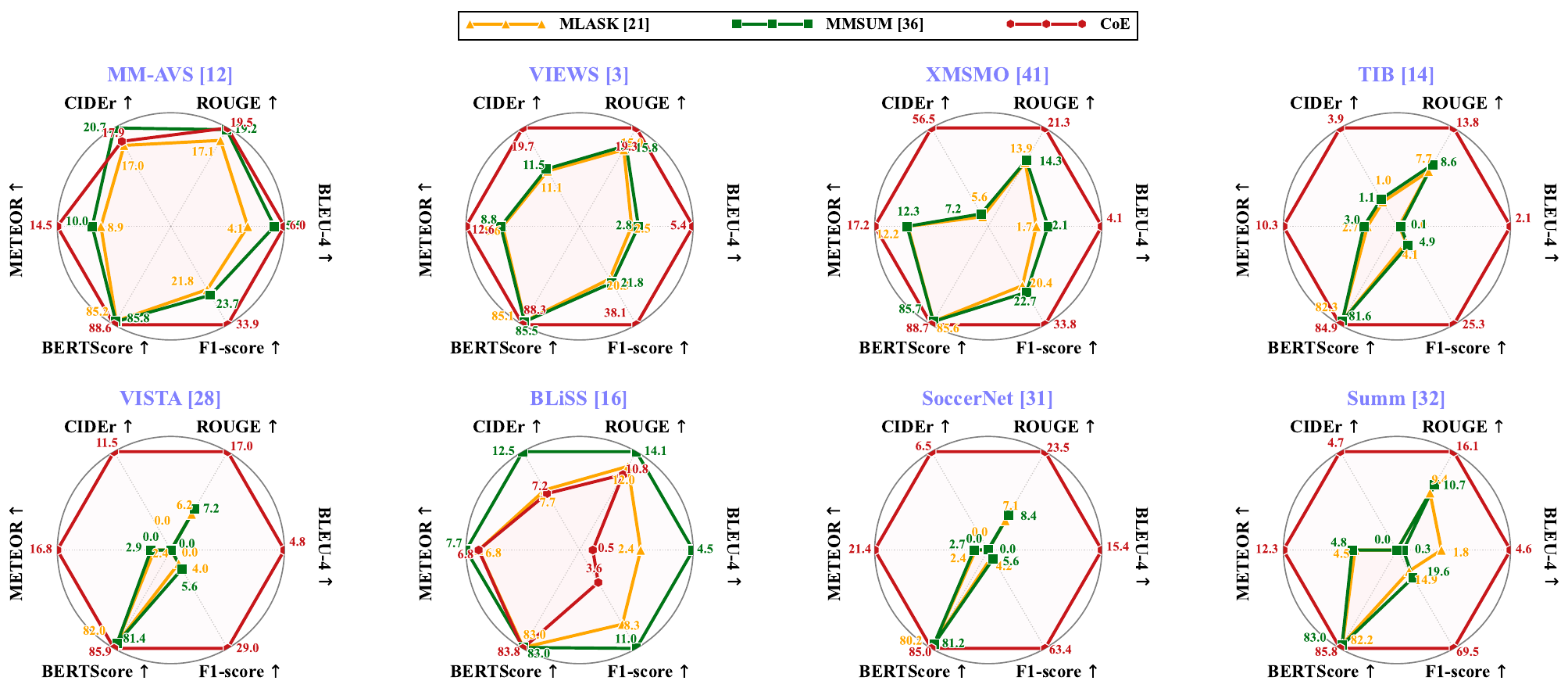}
  \vspace{-0.5em}
  \caption{\textbf{Motivating Experiments on MM-AVS.} Existing MMS models (e.g., \algo{MLASK}~\cite{krubinski2023mlask} and \algo{MMSum}~\cite{qiu2024mmsum}) achieve strong in-domain results when trained on MM-AVS~\cite{fu2021mm-avs}, but their performance drops sharply under domain shift. In contrast, our \textbf{training-free} \algo{CoE} framework generalizes effectively across diverse datasets, maintaining stable zero-shot performance without task-specific training or adaptation.}
  \label{fig:motivation-mmavs}
  \vspace{-0.5em}
\end{figure*}

\begin{figure*}[t]
  \centering
  \includegraphics[width=0.99\textwidth]{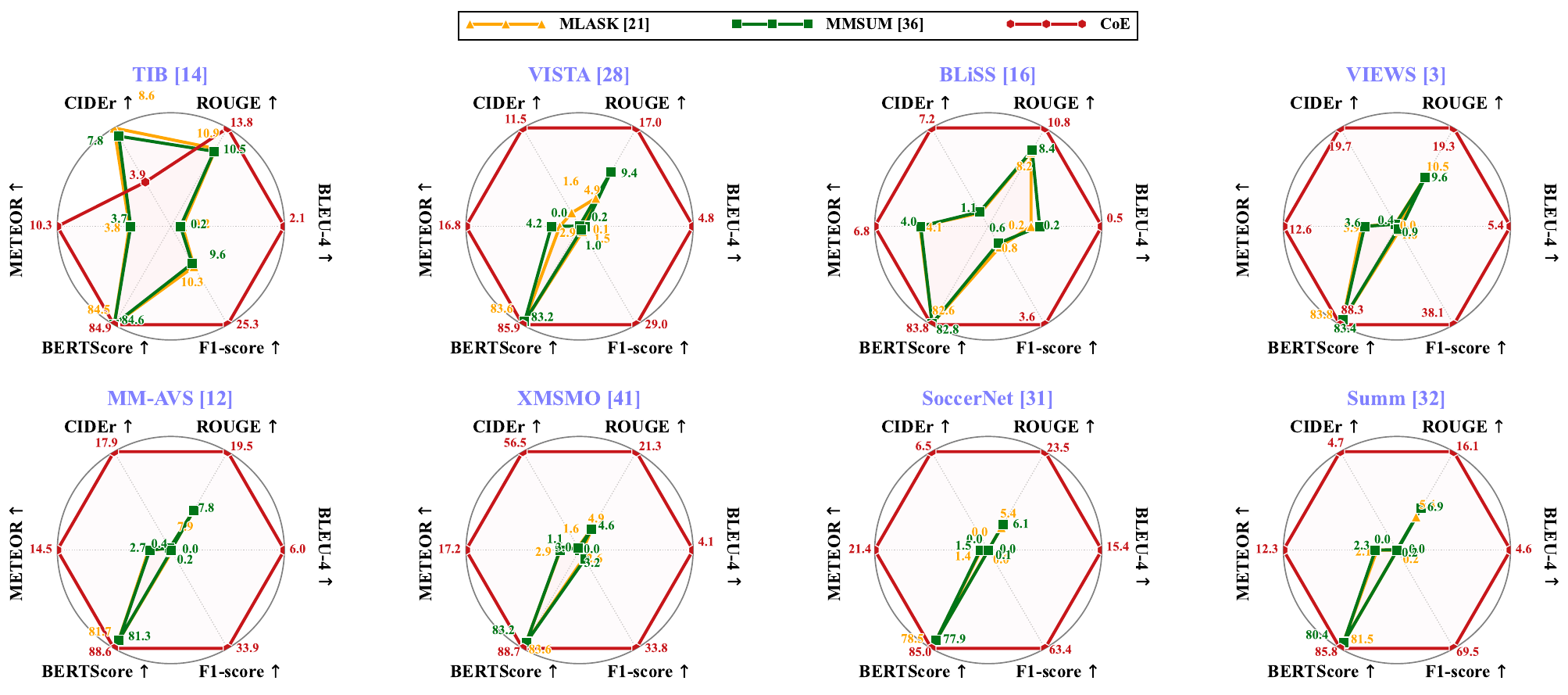}
  \vspace{-0.5em}
  \caption{\textbf{Motivating Experiments on TIB.} Existing MMS models (e.g., \algo{MLASK}~\cite{krubinski2023mlask} and \algo{MMSum}~\cite{qiu2024mmsum}) achieve strong in-domain results when trained on TIB~\cite{gigant2023tib}, but their performance drops sharply under domain shift. In contrast, our \textbf{training-free} \algo{CoE} framework generalizes effectively across diverse datasets, maintaining stable zero-shot performance without task-specific training or adaptation.}
  \label{fig:motivation-tib}
  \vspace{-0.5em}
\end{figure*}

\begin{figure*}[t]
  \centering
  \includegraphics[width=0.99\textwidth]{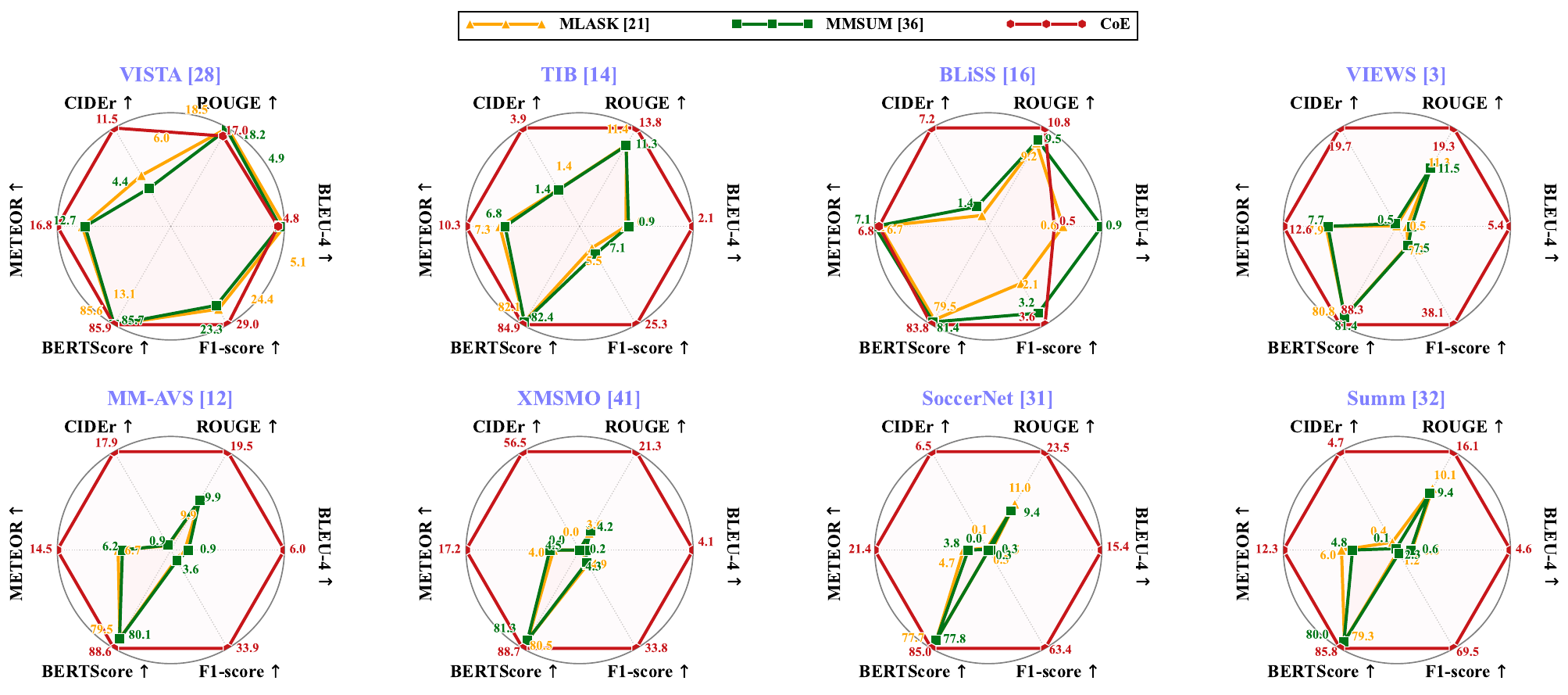}
  \vspace{-0.5em}
  \caption{\textbf{Motivating Experiments on VISTA.} Existing MMS models (e.g., \algo{MLASK}~\cite{krubinski2023mlask} and \algo{MMSum}~\cite{qiu2024mmsum}) achieve strong in-domain results when trained on VISTA~\cite{liu2025vista}, but their performance drops sharply under domain shift. In contrast, our \textbf{training-free} \algo{CoE} framework generalizes effectively across diverse datasets, maintaining stable zero-shot performance without task-specific training or adaptation.}
  \label{fig:motivation-vista}
  \vspace{-0.5em}
\end{figure*}

%%%Generalizability bliss

\begin{figure*}[t]
  \centering
  \includegraphics[width=0.99\textwidth]{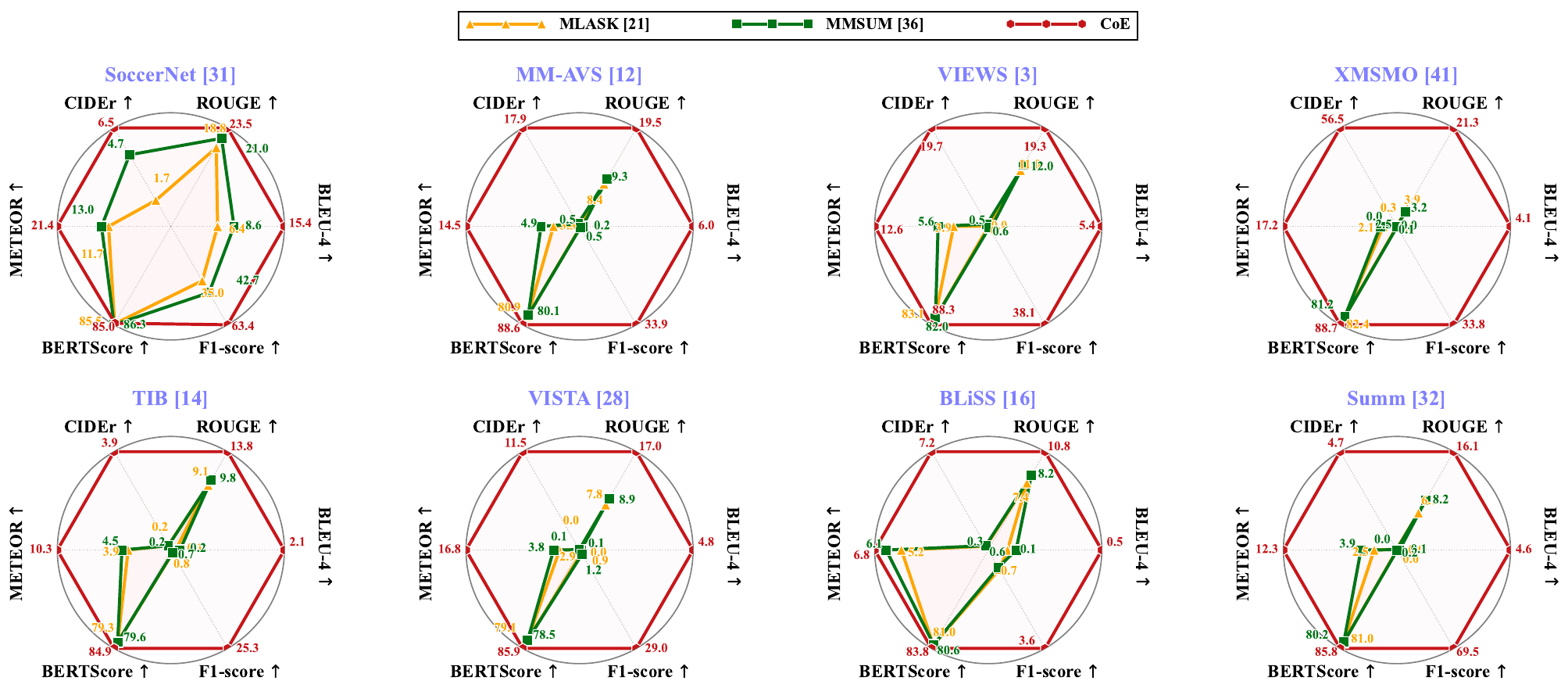}
  \vspace{-0.5em}
  \caption{\textbf{Motivating Experiments on SoccerNet.} Existing MMS models (e.g., \algo{MLASK}~\cite{krubinski2023mlask} and \algo{MMSum}~\cite{qiu2024mmsum}) achieve strong in-domain results when trained on SoccerNet~\cite{mkhallati2023soccernet}, but their performance drops sharply under domain shift. In contrast, our \textbf{training-free} \algo{CoE} framework generalizes effectively across diverse datasets, maintaining stable zero-shot performance without task-specific training or adaptation.}
  \label{fig:motivation-soccernet}
  \vspace{-0.5em}
\end{figure*}

\section{Dataset}
\label{sec:appen:dataset}
\input{tables/dataset}

To ensure a comprehensive evaluation across distinct video genres, we conduct experiments on eight diverse MMS benchmarks. 
%%%
These datasets cover five primary domains: news broadcasting, instructional videos, sports commentary, academic presentations, and entertainment narratives. 

This extensive selection allows us to validate the model's performance on varied content structures and linguistic styles. 
%%%
Specific details for each benchmark are described as follows:
\begin{itemize}
    \item \tb{VIEWS.} 
    Representing the news broadcasting domain, VIEWS~\cite{ayyubi2024views} is a large-scale benchmark originally designed for entity-aware video captioning. 
    %%%
    Derived from the $\text{M}^2\text{E}^2\text{R}$~\cite{ayyubi2022multimodal} corpus, it mitigates the loose alignment issues typical of news datasets by utilizing LLMs to generate ground-truth captions from shot-specific event descriptions. 
    %%%
    Unlike generic benchmarks, VIEWS requires models to explicitly identify named entities and interpret dynamic news contexts from visual cues. 
    %%%
    To adapt this dataset for the MMS task, we leverage the official video descriptions from YouTube to serve as the textual input.
    
    \item \tb{MM-AVS.}
    A full-scale benchmark in the news domain, MM-AVS~\cite{fu2021mm-avs} is curated from CNN and Daily Mail archives.
    %%%
    It distinguishes itself through comprehensive modality coverage, providing a rich set of inputs for each news story, including articles, videos, audio, transcripts, images, and captions. 
    %%%
    This dataset challenges models to synthesize heterogeneous information from these diverse streams into concise summaries, effectively addressing the lack of holistic audio-visual integration in prior benchmarks.

    \item \tb{XMSMO-News.} 
    Targeting the ``Too Long; Didn't Watch" (TL;DW) scenario, XMSMO-News (XMSMO)~\cite{tang2023tldw} establishes the task of \textit{extreme} MMS using content derived from BBC News. 
    %%%
    Unlike standard benchmarks, it challenges models to condense lengthy video-document pairs into an ultra-compact output comprising a single cover frame and a one-sentence summary. 
    %%%
    This rigorous constraint serves to evaluate the model's capacity to distill the most salient visual and textual semantics into a minimal representation, effectively addressing the challenge of information overload.

    \item \tb{TIB.}
    A benchmark for the academic domain, TIB~\cite{gigant2023tib} focuses on abstractive summarization of long-form videoconference recordings. 
    %%%
    Containing over 9,100 lectures with an average length of 37 minutes, it demands robust long-context modeling capabilities. 
    %%%
    Unlike standard datasets, TIB requires processing technically dense narratives and aligning asynchronous modalities—including speech, visual gestures, and presentation slides—to produce accurate summaries.

    \item \tb{VISTA.}    
    Centered on the domain of scientific communication, VISTA~\cite{liu2025vista} is a large-scale benchmark tailored for video-to-text summarization of academic discourse. 
    %%%
    It comprises 18,599 aligned pairs of conference recordings and their canonical paper abstracts, curated from premier AI venues such as the ACL Anthology, ICML, and NeurIPS. 
    %%%
    By targeting the gap in processing technical terminology and scientific visual elements, VISTA challenges models to synthesize long-form multimodal content (averaging 6.8 minutes) into highly structured, factually dense summaries.
    %%%
    To adapt this dataset for the MMS task, we leverage Whisper~\cite{radford2023whisper} to generate transcripts from the videos.

    \item \tb{BLiSS.}
    Sourced from the Behance platform, BLiSS~\cite{he2023bliss} serves as a premier benchmark for the livestreaming and instructional domains. 
    %%%
    It comprises over 13,000 video-transcript pairs specifically focused on creative artistic processes. 
    %%%
    Distinguished by its raw, untrimmed nature, BLiSS features content with extensive durations and slow-paced temporal dynamics, presenting a distinct challenge compared to traditional short-video datasets. 
    %%%
    Furthermore, it provides temporally aligned transcripts and dual-modality ground truth (key-frames and key-sentences), enabling rigorous evaluation of cross-modal alignment in long-horizon, realistic streaming environments.

    \item \tb{SoccerNet-Caption.}
    SoccerNet-Caption (SoccerNet)~\cite{mkhallati2023soccernet} is a large-scale benchmark specifically designed for dense video captioning within full-length soccer broadcasts.
    %%%
    It consists of 36,894 timestamped commentaries distributed across 715.9 hours of untrimmed video footage, sourced from 471 games in major European leagues.
    %%%
    Distinct from traditional benchmarks that rely on temporal intervals, this dataset introduces the Single-anchored Dense Video Captioning (SDVC) task, challenging models to generate rich, emotionally charged, and factually precise narratives anchored to specific game events.
    %%%
    With average video durations exceeding 45 minutes, it rigorously tests a model's ability to handle extensive temporal contexts and domain-specific terminology in a live broadcast setting.

    \item \tb{SummScreen$^\text{3D}$.}
    Curated for the entertainment narrative domain, SummScreen$^\text{3D}$ (Summ)~\cite{papalampidi2023hierarchical3d} constitutes a multimodal extension of the dialogue-centric SummScreen corpus~\cite{chen2022summscreen}.
    %%%
    It features 4,575 full-length TV episodes (averaging 40 minutes) paired with transcripts and abstractive summaries derived from soap operas.
    %%%
    This benchmark establishes a rigorous testbed for long-form video-to-text summarization, compelling models to synthesize visual and acoustic cues, such as character emotions and scene dynamics, with extensive dialogue to resolve long-range plot dependencies.
\end{itemize}

Detailed statistics regarding the training, validation, and testing splits for all datasets are summarized in Table~\ref{tab:datasets}.

\section{Implementation Details}
\label{sec:appen:imple}

\subsection{Implementation Details of Baselines}
To thoroughly evaluate the efficacy and superiority of our proposed \algo{CoE} framework, we employ two distinct classes of baseline models in our comparative experiments.
%%%
The first class comprises state-of-the-art traditional MMS methods, including MLASK~\cite{krubinski2023mlask} and MMSum~\cite{qiu2024mmsum}. These models are included primarily to examine the performance of our training-free \algo{CoE} framework under scenarios of domain transfer and to rigorously test its generalization capability, offering a crucial performance benchmark against domain-dependent counterparts.
%%%
The second class of baselines focuses on the video Chain-of-Thought (CoT) reasoning mechanisms, specifically encompassing TCoT~\cite{arnab2025tcot}, CoF~\cite{ghazanfari2025cof}, ViTCoT~\cite{zhang2025vitcot}, and CoS~\cite{hu2025cos}. This selection provides a direct comparison point, allowing us to quantify the superiority of the \algo{CoE} framework in structured event reasoning and event flow modeling over existing CoT strategies.
%%%
The implementation details are as follows:
\begin{itemize}
    \item \tb{MLASK.} 
    We fine-tune MLASK~\cite{krubinski2023mlask} using its official implementation. 
    %%%
    In addition to generating textual summaries, the model is designed to select a representative video frame as a cover image. 
    %%%
    Since our datasets do not contain ground-truth cover annotations, we construct the cover images as follows. 
    %%%
    For VIEWS, MM-AVS, XMSMO, SoccerNet, Summ, and BLiSS, we use CLIP~\cite{radford2021clip} to identify the frame with the highest semantic similarity to the transcript sentences and set it as the cover image. 
    %%%
    For TIB and VISTA, we directly use the first frame as the cover, as it typically corresponds to the title slide or agenda.
    
    \item \tb{MMSum.}
    We deploy MMSum~\cite{qiu2024mmsum} following its official implementation, strictly adhering to its standard protocols for feature extraction. 
    %%%
    Similar to MLASK, MMSum jointly generates a textual summary and a visual thumbnail.
    %%%
    To compensate for the lack of native thumbnail annotations in our datasets, we mirror the data processing strategy used for MLASK: we utilize the same CLIP-based alignment to derive pseudo-ground-truth thumbnails for general video domains, while defaulting to the initial frame as the target for presentation-oriented content (TIB and VISTA).
    
    \item \tb{TCoT.}
    Due to the absence of an official open-source implementation, we re-implement Temporal Chain of Thought (\algo{TCoT})~\cite{arnab2025tcot} strictly following the algorithmic design described in the original paper. 
    %%%
    To ensure a fair comparison, we instantiate the framework using the same MLLM backbone as our method (Qwen2.5-VL-7B-Instruct~\cite{bai2025qwen2}). 
    %%%
    Regarding hyperparameter settings, we adhere to the protocols reported in the original work: we adopt the \textit{Dynamic-Segment} inference strategy with a segment size of $s=64$ and a segment count of $l=12$. 
    %%%
    Input videos are sampled at 1 fps, and the maximum context window is constrained to 32K tokens to match the original experimental budget.

    \item \tb{CoF.}
    We use the official \texttt{CoF-8B} checkpoint from Chain-of-Frames (\algo{CoF})~\cite{ghazanfari2025cof}, which is fine-tuned on the InternVL3-8B~\cite{zhu2025internvl3} backbone. 
    %%%
    Regarding video processing, we uniformly sample frames at 1 fps and set the maximum input length to 64 frames. 
    %%%
    During inference, we adapt the original CoT prompt to the summarization task. 
    %%%
    Specifically, we modify the instruction to elicit frame-aware reasoning traces followed by a comprehensive video summary, rather than a question-specific answer.

    \item \tb{ViTCoT.}
    We re-implement Video-Text Interleaved Chain-of-Thought (\algo{ViTCoT})~\cite{zhang2025vitcot} on the same Qwen2.5-VL-7B-Instruct backbone as our method, strictly following the two-stage Video-Text Interleaved CoT paradigm and the prompt templates described in the original paper. 
    %%%
    In the first stage, the model performs standard text-only CoT reasoning conditioned on the video and transcript. 
    %%%
    In the second stage, we sample the raw video at 1 fps and use CLIP~\cite{radford2021clip} to select a small set of high-similarity frames, which are then interleaved into the intermediate reasoning trace to simulate the oracle key-video used in ViTCoT. 
    %%%
    The maximum number of key frames per video and other decoding hyperparameters follow the configuration reported by Zhang et al.~\cite{zhang2025vitcot}.

    \item \tb{CoS.}
    We adapt Chain-of-Shot prompting (\algo{CoS})~\cite{hu2025cos} to our MMS setting as a training-free baseline. 
    %%%
    Following the original design, we use an LLaVA-based~\cite{zhang2024videollava} classifier to perform binary video summarization over mosaiced frame groups, and LongVA as the downstream summarization backbone. 
    %%%
    For each video, we uniformly sample frames at 1 fps with at most 32 frames, group every four consecutive frames into a composite image for binary relevance prediction, and then construct positive and negative shot sequences from the resulting binary codes. 
    %%%
    The original QA-oriented prompts in CoS are rewritten into summarization-style instructions so that LongVA directly generates textual summaries conditioned on the selected shots, while other hyperparameters follow the settings in the original CoS paper.
\end{itemize}

%%% case study views
\begin{figure*}[t]
  \centering
  \begin{subfigure}[t]{0.49\textwidth}
    \centering
    \includegraphics[width=\linewidth]{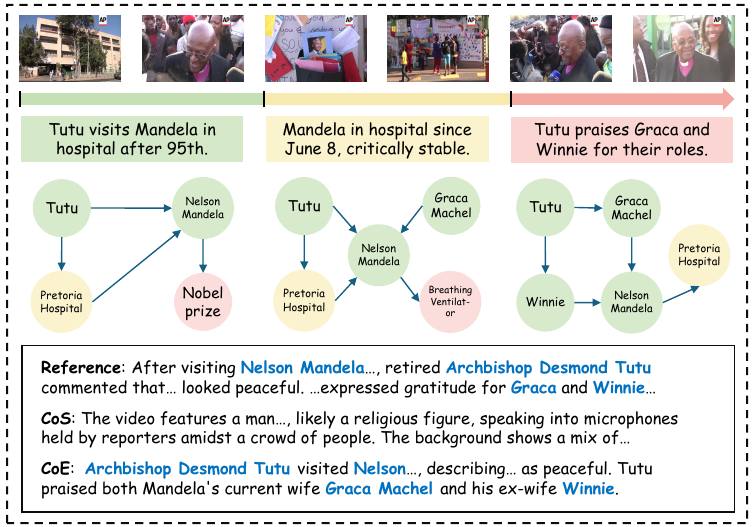}
    \caption{\algo{CoE} grounds news events mentions across video and news articles.}
    \label{fig:case_study_views_a}
  \end{subfigure}\hfill
  \begin{subfigure}[t]{0.49\textwidth}
    \centering
    \includegraphics[width=\linewidth]{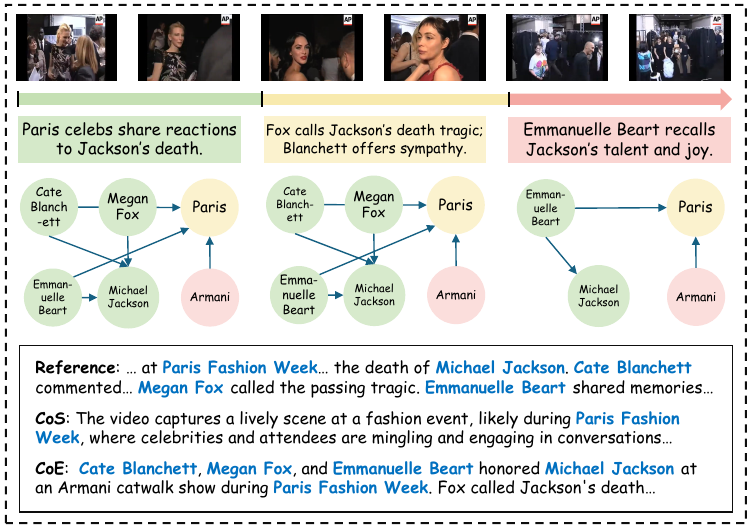}
    \caption{\algo{CoE} tracks event progression and produces concise news summaries.}
    \label{fig:case_study_views_b}
  \end{subfigure}
  \vspace{-0.5em}
  \caption{\tb{Case study on VIEWS}. 
  (a) Guided by the HEG, \algo{CoE} correctly links speeches by public figures to the corresponding entities and locations, aligning visual evidence with article-referenced mentions.
  (b) \algo{CoE} follows the evolution of the news story across scenes and generates compact news-style summaries, while the baseline mainly enumerates local visual details without capturing the full narrative.}
  \label{fig:case_study_views}
  \vspace{-0.5em}
\end{figure*}

%%% case study mmavs
\begin{figure*}[t]
  \centering
  \begin{subfigure}[t]{0.49\textwidth}
    \centering
    \includegraphics[width=\linewidth]{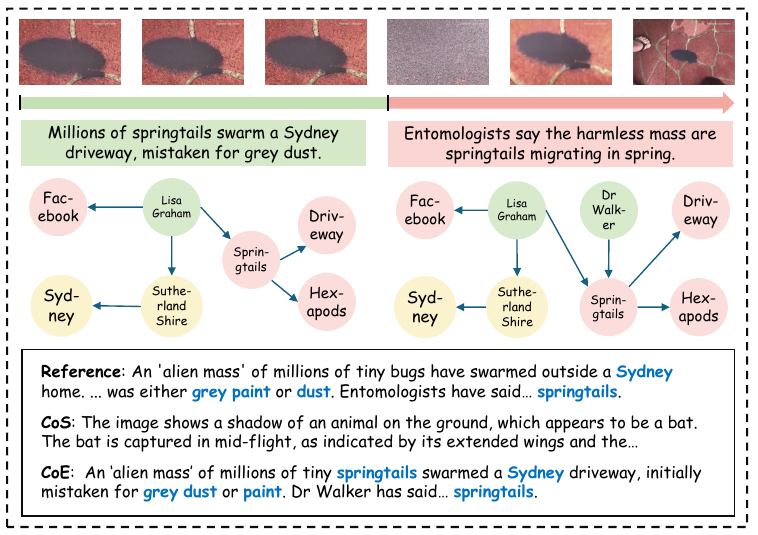}
    \caption{\algo{CoE} aligns news clips with the correct sub-events and entities.}
    \label{fig:case_study_mmavs_a}
  \end{subfigure}\hfill
  \begin{subfigure}[t]{0.49\textwidth}
    \centering
    \includegraphics[width=\linewidth]{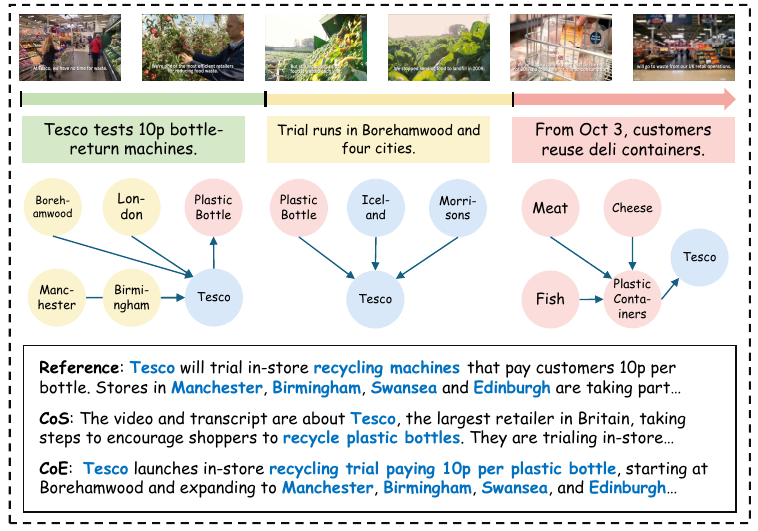}
    \caption{\algo{CoE} summarizes articles by following event development.}
    \label{fig:case_study_mmavs_b}
  \end{subfigure}
  \vspace{-0.5em}
  \caption{\tb{Case study on MM-AVS}. 
  (a) \algo{CoE} uses the event graph as a scaffold to attach each news clip to the correct sub-event and to ground entities such as locations, organizations, and key objects in the scene.
  (b) By aggregating grounded clips along the event trajectories, \algo{CoE} produces summaries that mirror how the news article develops, whereas the baseline remains close to scene-level description and often ignores the structure of the story.}
    \label{fig:case_study_mmavs}
  \vspace{-0.5em}
\end{figure*}

%%% case study xmsmo
\begin{figure*}[t]
  \centering
  \begin{subfigure}[t]{0.49\textwidth}
    \centering
    \includegraphics[width=\linewidth]{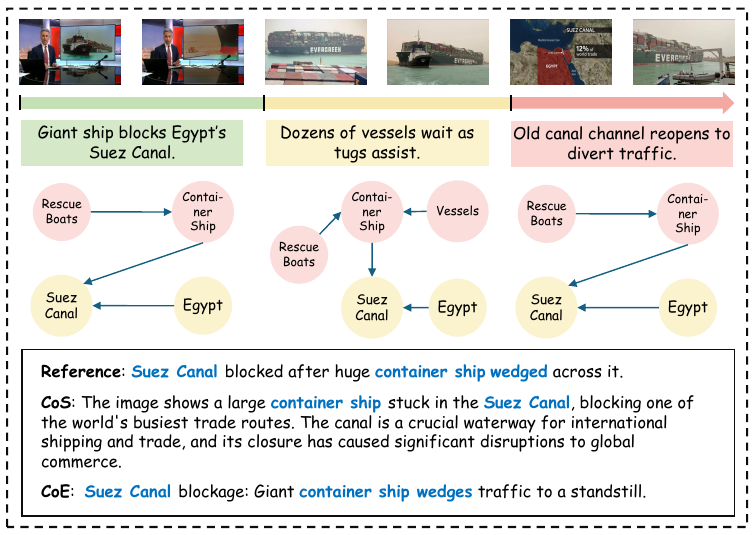}
    \caption{\algo{CoE} produces concise headline style summaries for extreme news cases.}
    \label{fig:case_study_xmsmo_a}
  \end{subfigure}\hfill
  \begin{subfigure}[t]{0.49\textwidth}
    \centering
    \includegraphics[width=\linewidth]{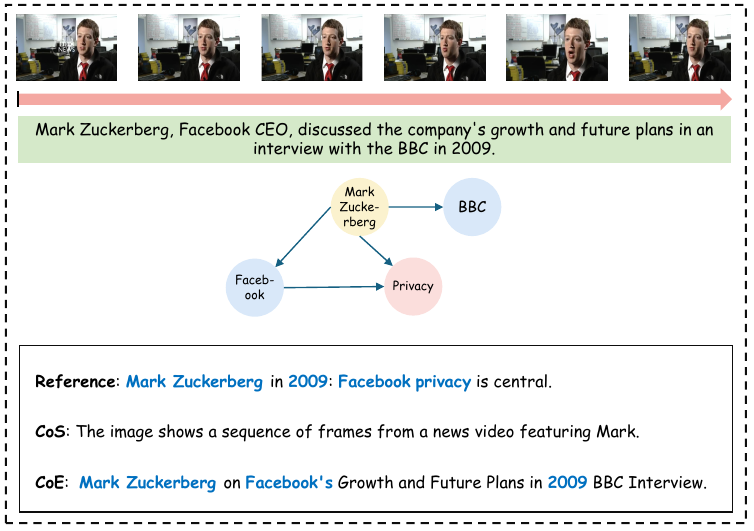}
    \caption{\algo{CoE} correctly identifies the main subject and interview context.}
    \label{fig:case_study_xmsmo_b}
  \end{subfigure}
  \vspace{-0.5em}
  \caption{\tb{Case study on XMSMO}. 
  (a) For the Suez Canal blockage example, \algo{CoE} compresses the multi-shot sequence into a short headline-style summary that closely follows the reference, while the baseline, which cannot exploit style exemplars, outputs a long verbose caption that reads like a literal description of the scene.
  (b) For the ``Mark Zuckerberg" interview, \algo{CoE} uses the hierarchical event graph and transcript to recover the correct identity and role of the speaker and to summarize the topic of the interview, whereas the baseline never names ``Zuckerberg" and only provides a generic description of a news video.}
    \label{fig:case_study_xmsmo}
  \vspace{-0.5em}
\end{figure*}

%%% case study tib
\begin{figure*}[t]
  \centering
  \begin{subfigure}[t]{0.49\textwidth}
    \centering
    \includegraphics[width=\linewidth]{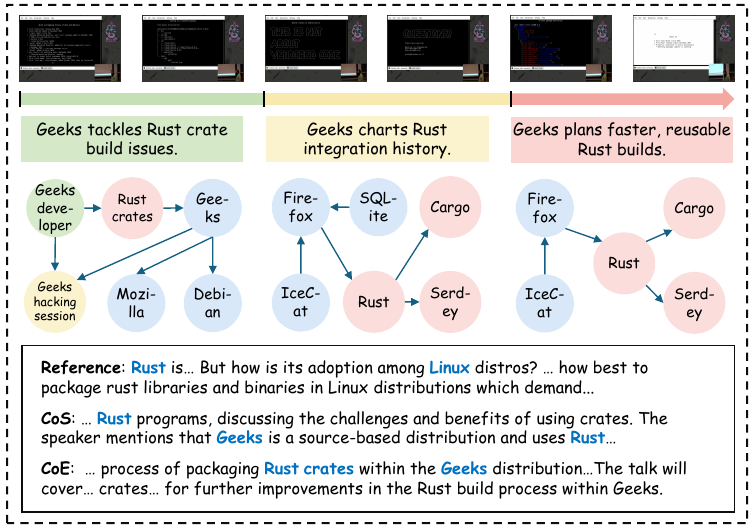}
    \caption{\algo{CoE} grounds technical concepts and entities in long lecture recordings.}
    \label{fig:case_study_tib_a}
  \end{subfigure}\hfill
  \begin{subfigure}[t]{0.49\textwidth}
    \centering
    \includegraphics[width=\linewidth]{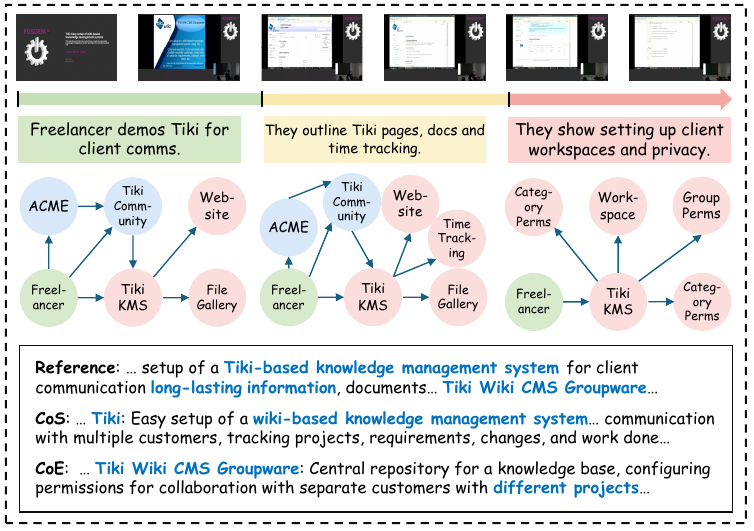}
    \caption{\algo{CoE} follows the structure of the talk and preserves key pedagogical steps.}
    \label{fig:case_study_tib_b}
  \end{subfigure}
  \vspace{-0.5em}
  \caption{\tb{Case study on TIB}. 
  (a) In technical lectures about topics such as Rust packaging or knowledge management with Tiki, \algo{CoE} aligns clips with the corresponding sub-events and grounds abstract entities like tools, libraries, and platforms using the slide and speaker context.
  (b) \algo{CoE} tracks how the speaker introduces the problem, presents solutions, and discusses future work, which leads to summaries that match the logical flow of the talk, whereas the baseline mainly reports isolated visual scenes without reconstructing the full argument.}
    \label{fig:case_study_tib}
  \vspace{-0.5em}
\end{figure*}

%%% case study VISTA
\begin{figure*}[t]
  \centering
  \begin{subfigure}[t]{0.49\textwidth}
    \centering
    \includegraphics[width=\linewidth]{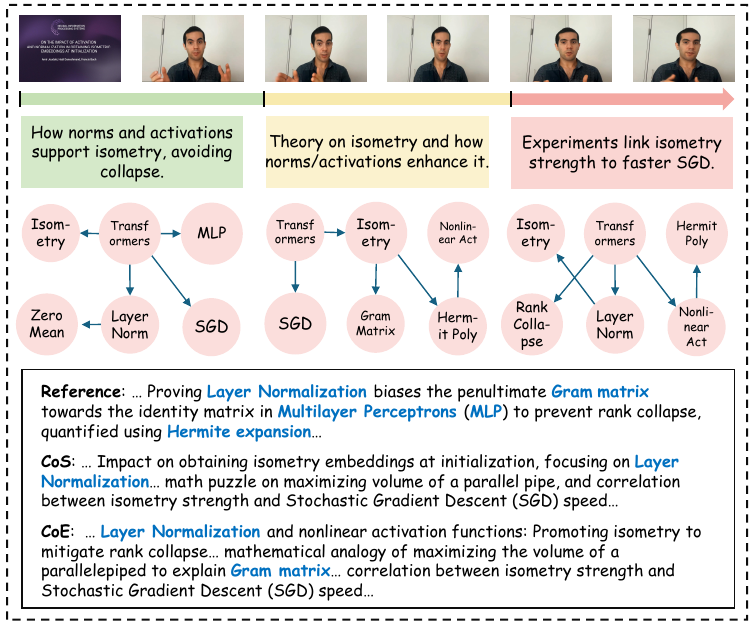}
    \caption{\algo{CoE} links slide content and spoken narration for technical talks.}
    \label{fig:case_study_vista_a}
  \end{subfigure}\hfill
  \begin{subfigure}[t]{0.49\textwidth}
    \centering
    \includegraphics[width=\linewidth]{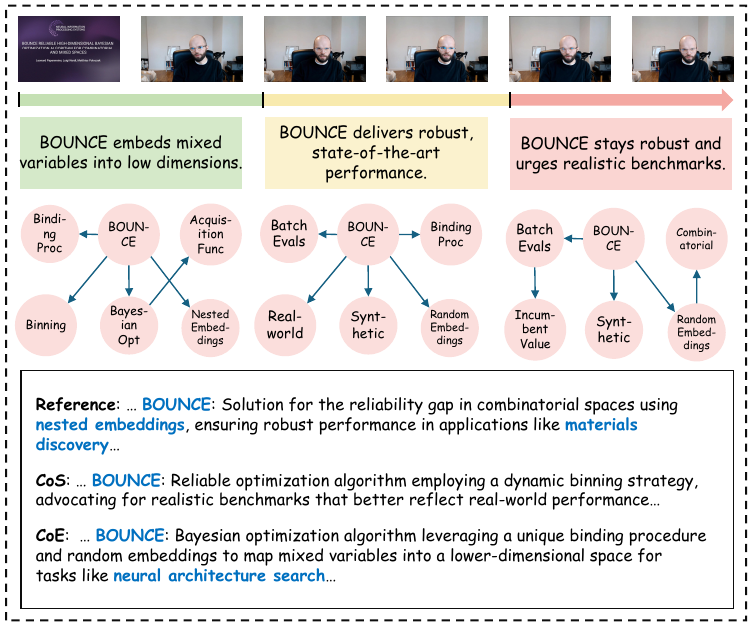}
    \caption{\algo{CoE} follows the progression from motivation to method and findings.}
    \label{fig:case_study_vista_b}
  \end{subfigure}
  \vspace{-0.5em}
  \caption{\tb{Case study on VISTA}. 
  (a) For a talk on isometry and layer normalization, \algo{CoE} uses the hierarchical event graph to segment the presentation into coherent conceptual blocks and to ground mathematical notions such as Gram matrix and stochastic gradient descent in the corresponding slide regions.
  (b) In the BOUNCE example, \algo{CoE} tracks how the speaker introduces the problem, presents the algorithm based on random embeddings for mixed variables, and summarizes empirical observations, which leads to a concise summary that highlights the main scientific message. \algo{CoS} also produces reasonable descriptions on VISTA, but its outputs are typically more generic and rely less on the detailed terminology that appears on the slides.}
    \label{fig:case_study_vista}
  \vspace{-0.5em}
\end{figure*}

\subsection{Implementation Details of CoE}
Our \algo{CoE} framework uses the Qwen2.5-VL-7B-Instruct as the backbone and 1 fps with a maximum of 72 frames. The frame segment size is set to 6 frames and max merged segments up to 5 (30 frames).
Our proposed \algo{CoE} framework is implemented based on the powerful Qwen2.5-VL-7B-Instruct, leveraging its strong multimodal understanding and long-context capabilities. 
%%%
As a training-free inference framework, the primary implementation configuration lies in efficient video sampling and structured prompt engineering.
%%%
Specifically, the input video is first uniformly sampled at a rate of 1 fps, and the maximum total number of frames is capped at 72. 
%%%
To facilitate the hierarchical event modeling of CoE, the sampled frames are grouped into frame segments, with each segment containing 6 frames. 
%%%
Subsequently, we constrain the prompt context to include a maximum of 5 merged segments (equating to a total of 30 frames), which are strategically interleaved with the Hierarchical Event Graph (HEG) and the textual transcript to guide the \algo{CoE} structured reasoning process. 
%%%
For decoding hyperparameters, the temperature is set to 0.1 and the maximum generated token count is fixed at 500.

\subsection{Implementation Details of Ablation Studies}
This subsection describes the implementation of ablations for the four modules in \algo{CoE}. 
%%%
Since the pipeline is progressive, with each stage consuming the structured outputs produced by its predecessor, we adopt a bypass strategy that disables one module at a time while keeping the remaining stages functional and directly comparable under identical inference settings. 

\paragraph{CoE -- HEG (w/o Hierarchical Event Graph Construction)}
HEG serves as a global-to-local semantic scaffold, organizing the narrative into a hierarchy from global events to sub-events and further to entity-relation subgraphs, which in turn informs subsequent grounding and reasoning.
%%%
To ablate HEG while keeping the downstream interfaces intact, we use the raw input text $T$ as the sole textual context. 
%%%
Specifically, we omit the construction of the three-level HEG and its associated sub-event graphs, and instead pass $T$ directly to later stages, where event and entity cues are extracted from $T$ whenever needed. 

\paragraph{CoE -- CSG (w/o Cross-modal Spatial Grounding)}
CSG associates each video clip with a corresponding sub-event anchor and grounds entity-relation triples by verifying their visual evidence, yielding visually supported subgraphs for subsequent reasoning.
%%%
In \textbf{CoE -- CSG}, we remove CSG while keeping the rest of the pipeline unchanged. Concretely, we assign the text-derived subgraph produced in HEG to each clip or segment directly, without performing clip-level entity identification or relation grounding. 
%%%
Consequently, EER and DSG operate on ungrounded, text-only subgraphs, thereby eliminating the benefit of fine-grained visual verification.

\paragraph{CoE -- EER (w/o Event Evolution Reasoning)}
EER groups semantically coherent clips into longer temporal segments and characterizes event trajectories by analyzing subgraph changes across adjacent segments.
In \textbf{CoE -- EER}, we disable EER and perform reasoning directly on the subgraphs matched in the previous stage, without modeling temporal evolution. Specifically, we do not compare subgraphs between neighboring segments or track the emergence and persistence of entities and relations over time. Instead, trajectory descriptions are generated solely from the subgraph(s) associated with the current clip or segment, removing explicit temporal transition modeling.

\paragraph{CoE -- DSG (w/o Domain-adaptive Summary Generation)}
DSG first synthesizes an event-centric summary from the inferred trajectories and then applies lightweight style adaptation using a small set of in-domain exemplars.
In \textbf{CoE -- DSG}, we retain the event-centric summary synthesis but omit the style adaptation step. The initial summary is directly taken as the final output without further rewriting, isolating the contribution of domain adaptation and assessing whether stylistic refinement is necessary for strong benchmark performance.

\section{Additional Experiments}
\label{sec:appen:expt}

\subsection{Case Study}
\label{sec:appen:case}
In the main paper, we provided a detailed qualitative analysis on the Summ dataset to illustrate the effectiveness of our hierarchical reasoning framework (Section~\ref{sec:expt:case}). 
%%%
To further examine the versatility and robustness of \algo{CoE} across heterogeneous domains, we include additional case studies on five benchmarks: VIEWS (Figure~\ref{fig:case_study_views}), MM-AVS (Figure~\ref{fig:case_study_mmavs}), XMSMO (Figure~\ref{fig:case_study_xmsmo}), TIB (Figure~\ref{fig:case_study_tib}), and VISTA (Figure~\ref{fig:case_study_vista}). 
%%%
These examples cover a broad range of video genres, including news broadcasts, instructional content, and scientific presentations.
%%%
Similar patterns can be observed from BLiSS~\cite{he2023bliss} and SoccerNet~\cite{mkhallati2023soccernet}. To be concise, we omit additional results here.

As shown in Figures~\ref{fig:case_study_views} through~\ref{fig:case_study_vista}, \algo{CoE} consistently recovers the correct hierarchical event structures, aligns sub-events with the corresponding visual evidence, and grounds entities and relations with fine-grained precision. 
%%%
Compared with baseline methods that tend to focus on local scene description, \algo{CoE} preserves global temporal consistency and accurately captures long-range narrative development, demonstrating the utility of hierarchical event modeling for diverse video styles and discourse forms.
%%%
On TIB and VISTA, we also observe that \algo{CoS}~\cite{hu2025cos} yields qualitatively reasonable summaries that are consistent with its quantitative trends on long video understanding and that it can accurately retrieve key visual details from extended sequences.

%%%
Overall, the case studies confirm that the event-centric design of \algo{CoE} generalizes effectively across domains without task-specific fine-tuning. 
%%%
The generated summaries remain grounded, coherent, and stylistically appropriate, further validating the advantages of explicit structured reasoning in MMS.

\begin{figure*}[t]
  \centering
  \includegraphics[width=0.99\textwidth]{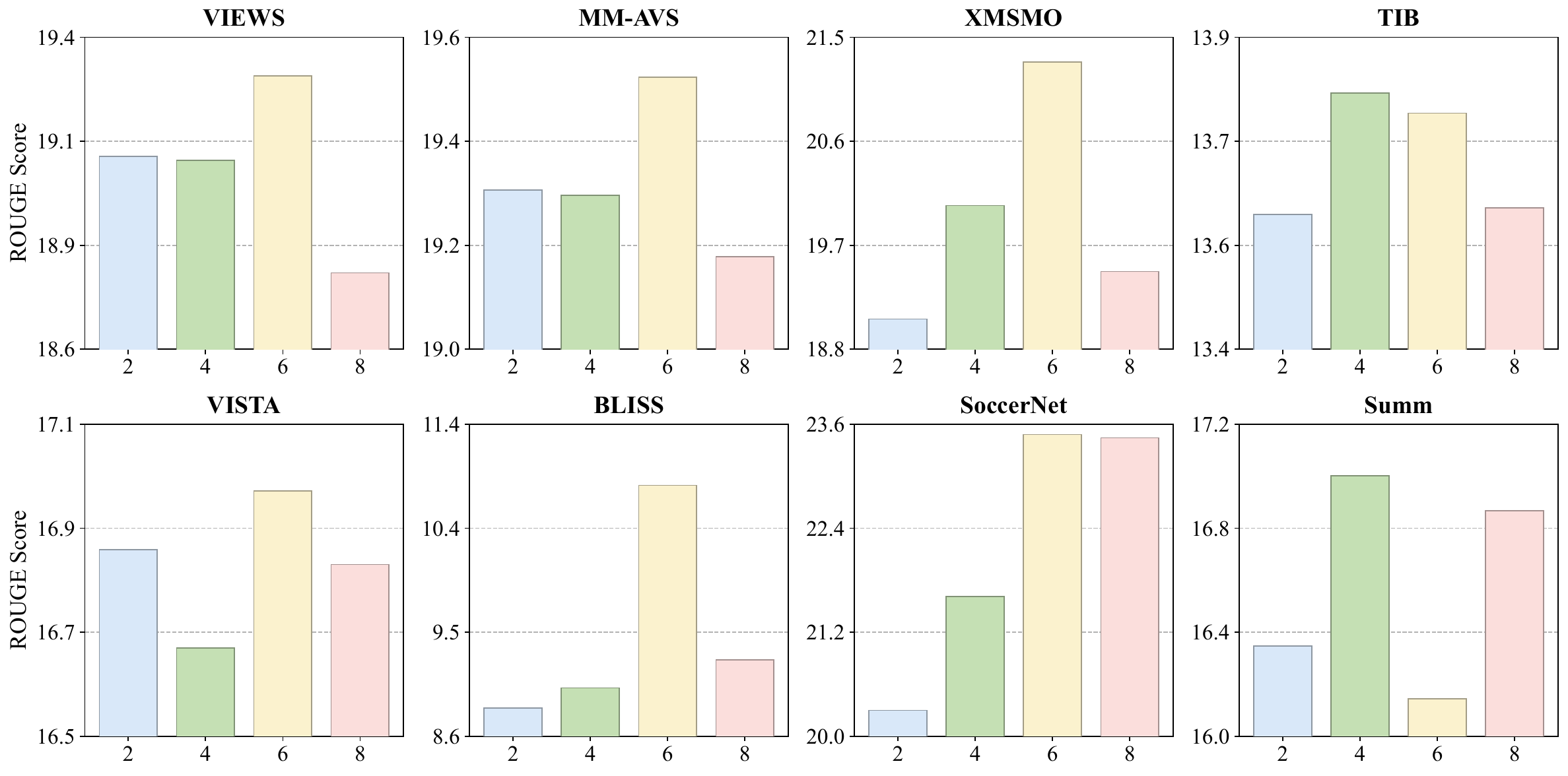}
  \vspace{-0.5em}
  \caption{\textbf{Effect of Temporal Granularity.} We report the ROUGE scores across eight datasets with varying video clip sizes ($K \in \{2, 4, 6, 8\}$). While performance varies slightly across domains, a clip size of $K=6$ yields the most robust and consistent improvements.}
  \label{fig:clip_size}
  % \vspace{-0.5em}
\end{figure*}

\subsection{Effect of Video Clip Size}
\label{sec:appen:clip}
To investigate the impact of temporal granularity on the reasoning capabilities of \algo{CoE}, we conduct an ablation study on the video clip size. 
%%%
Specifically, we fix the total number of sampled frames per video at 72 and vary the number of frames per clip, denoted as $K$, within the set $\{2, 4, 6, 8\}$. 
%%%
This setup alters the total number of temporal segments processed by the Event Evolution Reasoning (EER) module, directly influencing the model's ability to capture local motion dynamics versus global event transitions. 
%%%
By holding the total information budget constant, we isolate the effect of information distribution across temporal segments.

The results are illustrated in Figure~\ref{fig:clip_size}. 
%%%
We observe that a clip size of $K=6$ consistently achieves superior or competitive performance across the majority of benchmarks. 
%%%
Specifically, $K=6$ demonstrates a clear advantage on datasets requiring fine-grained motion understanding and long-term dependency modeling, such as XMSMO (21.28 ROUGE) and Soccernet (23.50 ROUGE). 
%%%
Furthermore, it yields the highest performance on BLiSS (10.83), significantly outperforming the $K=2$ baseline (8.85). 
%%%
While $K=8$ remains competitive in some scenarios, it does not consistently surpass the gains achieved by the 6-frame setting, and smaller clip sizes generally lag behind in performance metrics.

%%%
We attribute these findings to a trade-off between context sufficiency and information density.
%%%
Smaller clip sizes (e.g., $K=2$) tend to fragment the visual context, disrupting the continuity of actions and making it difficult for the model to form coherent event trajectories, as evidenced by suboptimal scores on TIB and VISTA. 
%%%
Conversely, excessively large clip sizes (e.g., $K=8$) may introduce visual redundancy or irrelevant background noise within a single processing unit, potentially diluting the salient features necessary for precise summarization. 
%%%
Therefore, we identify $K=6$ as the optimal balance point, providing sufficient temporal context for accurate entity-relation grounding without overwhelming the model with redundant visual information, and thus select it as the default setting for our \algo{CoE} framework.

\input{rebuttal_tables/time_efficiency}
\subsection{Inference Time}
\label{sec:appen:time}

To assess inference efficiency, we randomly sample 50 videos from each of the eight datasets and measure the end-to-end runtime per video under the same hardware and software settings for all methods. We report the average inference time across all sampled videos.
%%%
As shown in Table~\ref{tab:time-efficiency}, \algo{CoE} runs in \textbf{28.51}s per video on average, making it the second fastest method overall while maintaining strong overall performance in the main comparisons.

%% file: tables/dataset.tex
\begin{table}[t]
\centering
\small
\renewcommand{\arraystretch}{1.3}
\begin{tabular}{llrrr}
\toprule
\textbf{Dataset} & \textbf{Domain} & \textbf{\# Train} & \textbf{\# Val} & \textbf{\# Test} \\
\midrule
\cmidrule(lr){1-5}
VIEWS       & News              & 141.0 K & 1.6 K  & 1.6 K \\
MM-AVS      & News              & 1.8 K   & --     & 0.4 K \\
XMSMO       & News              & 4.4 K   & 0.3 K  & 0.3 K \\
\addlinespace[2pt]
\cmidrule(lr){1-5}
TIB         & Video Lectures    & 7.3 K   & 0.9 K  & 0.9 K \\
VISTA       & Academic          & 14.9 K  & 1.8 K  & 1.8 K \\
BLiSS       & Social Media      & 11.0 K  & 2.5 K  & 2.5 K \\
\addlinespace[2pt]
\cmidrule(lr){1-5}
SoccerNet   & Sports            & 0.5 K   & 0.06 K & 0.1 K \\
Summ        & Entertainment     & 5.2 K   & 3.0 K  & 3.0 K \\
\bottomrule
\end{tabular}
\caption{Summary of datasets used in our experiments, grouped by domain, along with the number of samples in the training, validation, and test splits.}
\label{tab:datasets}
\end{table}

%% file: rebuttal_tables/time_efficiency.tex
\begin{table}[t]
\centering
\small
\renewcommand{\arraystretch}{1.2}
% \setlength{\tabcolsep}{3pt}
% \vspace{-0.5em}
\resizebox{0.99\linewidth}{!}{
\begin{tabular}{cccccc}
  \toprule
  % \multirow{2.5}{*}{\tb{Metric}} & \multirow{2.5}{*}{\tb{Method}} & \multicolumn{8}{c}{\tb{Dataset}} \\
  % \cmidrule{3-10}
  % & & \tb{VIEWS} & \tb{MM-AVS} & \tb{XMSMO} & \tb{TIB} & \tb{VISTA} & \tb{BLiSS} & \tb{SoccerNet} & \tb{Summ} \\
  \textbf{Method} & \algo{TCoT} & \algo{CoF} & \algo{ViTCoT} & \algo{CoS} & \algo{CoE} \\
  \midrule
  Time (s) & 36.90 & 29.01 & 17.05 & 39.06 & 28.51 \\
  \bottomrule
\end{tabular}}
\caption{\textbf{Mean runtime comparison between \algo{CoE} and video CoT baselines.}}
\label{tab:time-efficiency}
\end{table}

%% file: main.bib
@String(CVPR= {IEEE Conf. Comput. Vis. Pattern Recog.})

@String(ICCV= {Int. Conf. Comput. Vis.})

@String(AAAI = {AAAI})

@String(CVPR  = {CVPR})

@String(ICCV  = {ICCV})

@InProceedings{fu2025videomme,
  author = {Fu, Chaoyou and Dai, Yuhan and Luo, Yongdong and Li, Lei and Ren, Shuhuai and Zhang, Renrui and Wang, Zihan and Zhou, Chenyu and Shen, Yunhang and Zhang, Mengdan and Chen, Peixian and Li, Yanwei and Lin, Shaohui and Zhao, Sirui and Li, Ke and Xu, Tong and Zheng, Xiawu and Chen, Enhong and Shan, Caifeng and He, Ran and Sun, Xing},
  title = {{Video-MME: The First-Ever Comprehensive Evaluation Benchmark of Multi-modal LLMs in Video Analysis}},
  booktitle = {Proceedings of the IEEE/CVF Conference on Computer Vision and Pattern Recognition (CVPR)},
  year = {2025},
  pages = {24108-24118}
}

@InProceedings{wang2025lvbench,
  author = {Wang, Weihan and He, Zehai and Hong, Wenyi and Cheng, Yean and Zhang, Xiaohan and Qi, Ji and Ding, Ming and Gu, Xiaotao and Huang, Shiyu and Xu, Bin and Dong, Yuxiao and Tang, Jie},
  title = {{LVBench: An Extreme Long Video Understanding Benchmark}},
  booktitle = {Proceedings of the IEEE/CVF International Conference on Computer Vision (ICCV)},
  year = {2025},
  pages = {22958-22967}
}

@InProceedings{li2024mvbench,
  author = {Li, Kunchang and Wang, Yali and He, Yinan and Li, Yizhuo and Wang, Yi and Liu, Yi and Wang, Zun and Xu, Jilan and Chen, Guo and Luo, Ping and Wang, Limin and Qiao, Yu},
  title = {{MVBench: A Comprehensive Multi-modal Video Understanding Benchmark}},
  booktitle = {Proceedings of the IEEE/CVF Conference on Computer Vision and Pattern Recognition (CVPR)},
  year = {2024},
  pages = {22195-22206}
}

@inproceedings{yu2019activitynet,
  title={{ActivityNet-QA:} a dataset for understanding complex web videos via question answering},
  author={Yu, Zhou and Xu, Dejing and Yu, Jun and Yu, Ting and Zhao, Zhou and Zhuang, Yueting and Tao, Dacheng},
  booktitle={Proceedings of the AAAI Conference on Artificial Intelligence},
  volume={33},
  number={01},
  pages={9127--9134},
  year={2019}
}

@inproceedings{han2025videoespresso,
  title={{VideoEspresso: A large-scale chain-of-thought dataset for fine-grained video reasoning via core frame selection}},
  author={Han, Songhao and Huang, Wei and Shi, Hairong and Zhuo, Le and Su, Xiu and Zhang, Shifeng and Zhou, Xu and Qi, Xiaojuan and Liao, Yue and Liu, Si},
  booktitle = {Proceedings of the IEEE/CVF Conference on Computer Vision and Pattern Recognition (CVPR)},
  pages={26181--26191},
  year={2025}
}

@inproceedings{zhang2025vitcot,
  title={{ViTCoT: Video-Text Interleaved Chain-of-Thought for Boosting Video Understanding in Large Language Models}},
  author={Zhang, Yongheng and Liu, Xu and Tao, Ruihan and Chen, Qiguang and Fei, Hao and Che, Wanxiang and Qin, Libo},
  booktitle={Proceedings of the 33rd ACM International Conference on Multimedia},
  pages={5267--5276},
  year={2025}
}

@article{wang2025cotasks,
  title={CoTasks: Chain-of-Thought based Video Instruction Tuning Tasks},
  author={Wang, Yanan and Vizcarra, Julio and Li, Zhi and Niu, Hao and Kurokawa, Mori},
  journal={arXiv preprint arXiv:2507.13609},
  year={2025}
}

@inproceedings{arnab2025tcot,
  title={Temporal Chain of Thought: Long-Video Understanding by Thinking in Frames},
  author={Anurag Arnab and Ahmet Iscen and Mathilde Caron and Alireza Fathi and Cordelia Schmid},
  booktitle={The Thirty-ninth Annual Conference on Neural Information Processing Systems (NeurIPS)},
  year={2025},
  url={https://openreview.net/forum?id=BDkNRlGmP9}
}

@inproceedings{ghazanfari2025cof,
  title={Chain-of-Frames: Advancing Video Understanding in Multimodal {LLM}s via Frame-Aware Reasoning},
  author={Sara Ghazanfari and Francesco Croce and Nicolas Flammarion and Prashanth Krishnamurthy and Farshad Khorrami and Siddharth Garg},
  booktitle={Proceedings of the 39th Annual Conference on Neural Information Processing Systems 2025 Workshop: First Workshop on Foundations of Reasoning in Language Models},
  year={2025},
  url={https://openreview.net/forum?id=8oGG6c0WFd}
}

@inproceedings{lee2025video-skill,
  title = "Video-Skill-{C}o{T}: Skill-based Chain-of-Thoughts for Domain-Adaptive Video Reasoning",
  author = "Lee, Daeun and Yoon, Jaehong and Cho, Jaemin and Bansal, Mohit",
  booktitle = "Findings of the Association for Computational Linguistics: EMNLP 2025",
  month = nov,
  year = "2025",
  pages = "18435--18449"
}

@article{wang2025videochat-a1,
  title={{VideoChat-A1: Thinking with Long Videos by Chain-of-Shot Reasoning}},
  author={Wang, Zikang and Chen, Boyu and Yue, Zhengrong and Wang, Yi and Qiao, Yu and Wang, Limin and Wang, Yali},
  journal={arXiv preprint arXiv:2506.06097},
  year={2025}
}

@article{zhang2026silvr,
title={Si{LVR}: A Simple Language-based Video Reasoning Framework},
author={Ce Zhang and Yan-Bo Lin and Ziyang Wang and Mohit Bansal and Gedas Bertasius},
journal={Transactions on Machine Learning Research (TMLR)},
issn={2835-8856},
year={2026}
}

@article{cheng2025vstar,
  title={{V-STaR: Benchmarking Video-LLMs on Video Spatio-Temporal Reasoning}},
  author={Cheng, Zixu and Hu, Jian and Liu, Ziquan and Si, Chenyang and Li, Wei and Gong, Shaogang},
  journal={arXiv preprint arXiv:2503.11495},
  year={2025}
}

@article{hu2025cos,
  title={{CoS: Chain-of-Shot Prompting for Long Video Understanding}},
  author={Hu, Jian and Cheng, Zixu and Si, Chenyang and Li, Wei and Gong, Shaogang},
  journal={arXiv preprint arXiv:2502.06428},
  year={2025}
}

@inproceedings{tang2025cardiff,
  title={{CaRDiff: Video Salient Object Ranking Chain of Thought Reasoning for Saliency Prediction with Diffusion}},
  author={Tang, Yunlong and Zhan, Gen and Yang, Li and Liao, Yiting and Xu, Chenliang},
  booktitle={Proceedings of the AAAI Conference on Artificial Intelligence},
  volume={39},
  number={7},
  pages={7302--7310},
  year={2025}
}

@inproceedings{liao2024videoinsta,
  title={{VideoINSTA: Zero-shot Long Video Understanding via Informative Spatial-Temporal Reasoning with LLMs}},
  author={Liao, Ruotong and Erler, Max and Wang, Huiyu and Zhai, Guangyao and Zhang, Gengyuan and Ma, Yunpu and Tresp, Volker},
  booktitle={Findings of the Association for Computational Linguistics: EMNLP 2024},
  pages={6577--6602},
  year={2024}
}

@inproceedings{fei2024vot,
  title={{Video-of-Thought: Step-by-Step Video Reasoning from Perception to Cognition}},
  author={Fei, Hao and Wu, Shengqiong and Ji, Wei and Zhang, Hanwang and Zhang, Meishan and Lee, Mong Li and Hsu, Wynne},
  booktitle={Proceedings of the 41st International Conference on Machine Learning (ICML)},
  pages={13109--13125},
  year={2024}
}

@inproceedings{wang2024videocot,
  title={{VideoCoT: A Video Chain-of-Thought Dataset with Active Annotation Tool}},
  author={Wang, Yan and Zeng, Yawen and Zheng, Jingsheng and Xing, Xiaofen and Xu, Jin and Xu, Xiangmin},
  booktitle={Proceedings of the 3rd Workshop on Advances in Language and Vision Research (ALVR)},
  pages={92--101},
  year={2024}
}

@inproceedings{jin2024graphcot,
  title={{Graph Chain-of-Thought: Augmenting Large Language Models by Reasoning on Graphs}},
  author={Jin, Bowen and Xie, Chulin and Zhang, Jiawei and Roy, Kashob Kumar and Zhang, Yu and Li, Zheng and Li, Ruirui and Tang, Xianfeng and Wang, Suhang and Meng, Yu and others},
  booktitle={Findings of the Association for Computational Linguistics ACL 2024},
  pages={163--184},
  year={2024}
}

@article{wu2023role,
  title={{The Role of Chain-of-Thought in Complex Vision-Language Reasoning Task}},
  author={Wu, Yifan and Zhang, Pengchuan and Xiong, Wenhan and Oguz, Barlas and Gee, James C and Nie, Yixin},
  journal={arXiv preprint arXiv:2311.09193},
  year={2023}
}

@inproceedings{wang2023plan,
  title={{Plan-and-Solve Prompting: Improving Zero-Shot Chain-of-Thought Reasoning by Large Language Models}},
  author={Wang, Lei and Xu, Wanyu and Lan, Yihuai and Hu, Zhiqiang and Lan, Yunshi and Lee, Roy Ka-Wei and Lim, Ee-Peng},
  booktitle={Proceedings of the 61st Annual Meeting of the Association for Computational Linguistics (Volume 1: Long Papers)},
  pages={2609--2634},
  year={2023}
}

@inproceedings{wei2022chain,
  title={{Chain-of-Thought Prompting Elicits Reasoning in Large Language Models}},
  author={Wei, Jason and Wang, Xuezhi and Schuurmans, Dale and Bosma, Maarten and Ichter, Brian and Xia, Fei and Chi, Ed H and Le, Quoc V and Zhou, Denny},
  booktitle={Proceedings of the 36th International Conference on Neural Information Processing Systems (NeurIPS)},
  pages={24824--24837},
  year={2022}
}

@article{beedu2025hiersum,
  title={{HierSum: A Global and Local Attention Mechanism for Video Summarization}},
  author={Beedu, Apoorva and Essa, Irfan},
  journal={arXiv preprint arXiv:2504.18689},
  year={2025}
}

@inproceedings{tan2025smsmo,
  title={{Enhancing Large Language Models for Scientific Multimodal Summarization with Multimodal Output}},
  author={Tan, Zusheng and Zhong, Xinyi and Ji, Jing-Yu and Jiang, Wei and Chiu, Billy},
  booktitle={Proceedings of the 31st International Conference on Computational Linguistics: Industry Track},
  pages={263--275},
  year={2025}
}

@article{ali2025systematic,
  title={{A Systematic Literature Review on Multimodal Text Summarization}},
  author={Ali, Abid and Molla, Diego},
  journal={ACM Computing Surveys},
  volume={58},
  number={3},
  pages={1--38},
  year={2025}
}

@inproceedings{liu2025vista,
  title={{What Is That Talk About? A Video-to-Text Summarization Dataset for Scientific Presentations}},
  author={Liu, Dongqi and Whitehouse, Chenxi and Yu, Xi and Mahon, Louis and Saxena, Rohit and Zhao, Zheng and Qiu, Yifu and Lapata, Mirella and Demberg, Vera},
  booktitle={Proceedings of the 63rd Annual Meeting of the Association for Computational Linguistics (Volume 1: Long Papers)},
  pages={6187--6210},
  year={2025}
}

@article{liu2024multimodal,
  title={{Multimodal Cross-Lingual Summarization for Videos: A Revisit in Knowledge Distillation Induced Triple-Stage Training Method}},
  author={Liu, Nayu and Wei, Kaiwen and Yang, Yong and Tao, Jianhua and Sun, Xian and Yao, Fanglong and Yu, Hongfeng and Jin, Li and Lv, Zhao and Fan, Cunhang},
  journal={IEEE Transactions on Pattern Analysis and Machine Intelligence},
  volume={46},
  number={12},
  pages={10697--10714},
  year={2024}
}

@inproceedings{liu2024sitransformer,
  title={{Sitransformer: Shared information-guided transformer for extreme multimodal summarization}},
  author={Liu, Sicheng and Wang, Lintao and Zhu, Xiaogang and Lu, Xuequan and Wang, Zhiyong and Hu, Kun},
  booktitle={Proceedings of the 6th ACM International Conference on Multimedia in Asia},
  pages={1--7},
  year={2024}
}

@inproceedings{faheem2024urdumasd,
  title={{UrduMASD: a Multimodal Abstractive Summarization Dataset for Urdu}},
  author={Faheem, Ali and Ullah, Faizad and Ayub, Muhammad Sohaib and Karim, Asim},
  booktitle={Proceedings of the 2024 Joint International Conference on Computational Linguistics, Language Resources and Evaluation (LREC-COLING 2024)},
  pages={17245--17253},
  year={2024}
}

@inproceedings{ayyubi2024views,
  title={{VIEWS: Entity-Aware News Video Captioning}},
  author={Ayyubi, Hammad and Liu, Tianqi and Nagrani, Arsha and Lin, Xudong and Zhang, Mingda and Arnab, Anurag and Han, Feng and Zhu, Yukun and Feng, Xuande and Zhang, Kevin and others},
  booktitle={Proceedings of the 2024 Conference on Empirical Methods in Natural Language Processing (EMNLP)},
  pages={20220--20239},
  year={2024}
}

@inproceedings{mkhallati2023soccernet,
  title={{SoccerNet-caption: Dense video captioning for soccer broadcasts commentaries}},
  author={Mkhallati, Hassan and Cioppa, Anthony and Giancola, Silvio and Ghanem, Bernard and Van Droogenbroeck, Marc},
  booktitle={Proceedings of the IEEE/CVF Conference on Computer Vision and Pattern Recognition (CVPR)},
  pages={5074--5085},
  year={2023}
}

@inproceedings{qiu2024mmsum,
  title={{MMSum: A Dataset for Multimodal Summarization and Thumbnail Generation of Videos}},
  author={Qiu, Jielin and Zhu, Jiacheng and Han, William and Kumar, Aditesh and Mittal, Karthik and Jin, Claire and Yang, Zhengyuan and Li, Linjie and Wang, Jianfeng and Zhao, Ding and others},
  booktitle={Proceedings of the IEEE/CVF Conference on Computer Vision and Pattern Recognition (CVPR)},
  pages={21909--21921},
  year={2024}
}

@inproceedings{papalampidi2023hierarchical3d,
  title={{Hierarchical3D Adapters for Long Video-to-text Summarization}},
  author={Papalampidi, Pinelopi and Lapata, Mirella},
  booktitle={Findings of the Association for Computational Linguistics: EACL 2023},
  pages={1297--1320},
  year={2023}
}

@inproceedings{he2023bliss,
  title={{Align and Attend: Multimodal Summarization with Dual Contrastive Losses}},
  author={He, Bo and Wang, Jun and Qiu, Jielin and Bui, Trung and Shrivastava, Abhinav and Wang, Zhaowen},
  booktitle={Proceedings of the IEEE/CVF Conference on Computer Vision and Pattern Recognition (CVPR)},
  pages={14867--14878},
  year={2023}
}

@inproceedings{gigant2023tib,
  title={{TIB: A Dataset for Abstractive Summarization of Long Multimodal Videoconference Records}},
  author={Gigant, Th{\'e}o and Dufaux, Fr{\'e}d{\'e}ric and Guinaudeau, Camille and Decombas, Marc},
  booktitle={Proceedings of the 20th International Conference on Content-based Multimedia Indexing},
  pages={61--70},
  year={2023}
}

@inproceedings{krubinski2023mlask,
  title={{MLASK: Multimodal Summarization of Video-based News Articles}},
  author={Krubi{\'n}ski, Mateusz and Pecina, Pavel},
  booktitle={Findings of the association for computational linguistics: EACL 2023},
  pages={910--924},
  year={2023}
}

@article{tang2023tldw,
  title={{TLDW: Extreme Multimodal Summarization of News Videos}},
  author={Tang, Peggy and Hu, Kun and Zhang, Lei and Luo, Jiebo and Wang, Zhiyong},
  journal={IEEE Transactions on Circuits and Systems for Video Technology},
  volume={34},
  number={3},
  pages={1469--1480},
  year={2023}
}

@inproceedings{qiu2023sccs,
  title={{SCCS: Semantics-Consistent Cross-domain Summarization via Optimal Transport Alignment}},
  author={Qiu, Jielin and Zhu, Jiacheng and Xu, Mengdi and Dernoncourt, Franck and Bui, Trung and Wang, Zhaowen and Li, Bo and Zhao, Ding and Jin, Hailin},
  booktitle={Findings of the Association for Computational Linguistics: ACL 2023},
  pages={1584--1601},
  year={2023}
}

@inproceedings{chen2022summscreen,
  title={Summscreen: A dataset for abstractive screenplay summarization},
  author={Chen, Mingda and Chu, Zewei and Wiseman, Sam and Gimpel, Kevin},
  booktitle={Proceedings of the 60th Annual Meeting of the Association for Computational Linguistics (Volume 1: Long Papers)},
  pages={8602--8615},
  year={2022}
}

@article{ayyubi2022multimodal,
  title={Multimodal event graphs: Towards event centric understanding of multimodal world},
  author={Ayyubi, Hammad A and Thomas, Christopher and Chum, Lovish and Lokesh, Rahul and Niu, Yulei and Lin, Xudong and Chen, Long and Koo, Jaywon and Ray, Sounak and Chang, Shih-Fu},
  journal={arXiv preprint arXiv:2206.07207},
  year={2022}
}

@inproceedings{fu2021mm-avs,
  title={{MM-AVS: A Full-Scale Dataset for Multi-modal Summarization}},
  author={Fu, Xiyan and Wang, Jun and Yang, Zhenglu},
  booktitle={Proceedings of the 2021 Conference of the North American Chapter of the Association for Computational Linguistics: Human Language Technologies},
  pages={5922--5926},
  year={2021}
}

@inproceedings{li2020vmsmo,
  title={{VMSMO: Learning to Generate Multimodal Summary for Video-based News Articles}},
  author={Li, Mingzhe and Chen, Xiuying and Gao, Shen and Chan, Zhangming and Zhao, Dongyan and Yan, Rui},
  booktitle={Proceedings of the 2020 Conference on Empirical Methods in Natural Language Processing (EMNLP)},
  pages={9360--9369},
  year={2020}
}

@inproceedings{sanabria18how2,
  title = {{How2:} A Large-scale Dataset For Multimodal Language Understanding},
  author = {Sanabria, Ramon and Caglayan, Ozan and Palaskar, Shruti and Elliott, Desmond and Barrault, Lo\"ic and Specia, Lucia and Metze, Florian},
  booktitle = {Proceedings of the Workshop on Visually Grounded Interaction and Language (ViGIL)},
  year = {2018},
  organization={NeurIPS},
  url = {http://arxiv.org/abs/1811.00347}
}

@article{bai2025qwen2,
  title={{Qwen2.5-VL Technical Report}},
  author={Shuai Bai and Keqin Chen and Xuejing Liu and Jialin Wang and Wenbin Ge and Sibo Song and Kai Dang and Peng Wang and Shijie Wang and Jun Tang and Humen Zhong and Yuanzhi Zhu and Mingkun Yang and Zhaohai Li and Jianqiang Wan and Pengfei Wang and Wei Ding and Zheren Fu and Yiheng Xu and Jiabo Ye and Xi Zhang and Tianbao Xie and Zesen Cheng and Hang Zhang and Zhibo Yang and Haiyang Xu and Junyang Lin},
  journal={arXiv preprint arXiv:2502.13923},
  year={2025}
}

@article{comanici2025gemini,
  title={{Gemini 2.5: Pushing the Frontier with Advanced Reasoning, Multimodality, Long Context, and Next Generation Agentic Capabilities}},
  author={Comanici, Gheorghe and Bieber, Eric and Schaekermann, Mike and Pasupat, Ice and Sachdeva, Noveen and Dhillon, Inderjit and Blistein, Marcel and Ram, Ori and Zhang, Dan and Rosen, Evan and others},
  journal={arXiv preprint arXiv:2507.06261},
  year={2025}
}

@article{zhu2025internvl3,
  title={Internvl3: Exploring advanced training and test-time recipes for open-source multimodal models},
  author={Zhu, Jinguo and Wang, Weiyun and Chen, Zhe and Liu, Zhaoyang and Ye, Shenglong and Gu, Lixin and Tian, Hao and Duan, Yuchen and Su, Weijie and Shao, Jie and others},
  journal={arXiv preprint arXiv:2504.10479},
  year={2025}
}

@article{zhang2024videollava,
  title={Video instruction tuning with synthetic data},
  author={Zhang, Yuanhan and Wu, Jinming and Li, Wei and Li, Bo and Ma, Zejun and Liu, Ziwei and Li, Chunyuan},
  journal={arXiv preprint arXiv:2410.02713},
  year={2024}
}

@article{zhang2025long,
  title={Long Context Transfer from Language to Vision},
  author={Peiyuan Zhang and Kaichen Zhang and Bo Li and Guangtao Zeng and Jingkang Yang and Yuanhan Zhang and Ziyue Wang and Haoran Tan and Chunyuan Li and Ziwei Liu},
  journal={Transactions on Machine Learning Research (TMLR)},
  issn={2835-8856},
  year={2025},
  url={https://openreview.net/forum?id=30RAWQVGlx},
}

@article{hurst2024gpt,
  title={{GPT-4o System Card}},
  author={Hurst, Aaron and Lerer, Adam and Goucher, Adam P and Perelman, Adam and Ramesh, Aditya and Clark, Aidan and Ostrow, AJ and Welihinda, Akila and Hayes, Alan and Radford, Alec and others},
  journal={arXiv preprint arXiv:2410.21276},
  year={2024}
}

@inproceedings{radford2023whisper,
  title={{Robust Speech Recognition via Large-Scale Weak Supervision}},
  author={Radford, Alec and Kim, Jong Wook and Xu, Tao and Brockman, Greg and McLeavey, Christine and Sutskever, Ilya},
  booktitle={Proceedings of the 40th International Conference on Machine Learning},
  pages={28492--28518},
  year={2023}
}

@inproceedings{kwon2023vllm,
  title={Efficient memory management for large language model serving with pagedattention},
  author={Kwon, Woosuk and Li, Zhuohan and Zhuang, Siyuan and Sheng, Ying and Zheng, Lianmin and Yu, Cody Hao and Gonzalez, Joseph and Zhang, Hao and Stoica, Ion},
  booktitle={Proceedings of the 29th symposium on operating systems principles},
  pages={611--626},
  year={2023}
}

@inproceedings{radford2021clip,
  title={Learning transferable visual models from natural language supervision},
  author={Radford, Alec and Kim, Jong Wook and Hallacy, Chris and Ramesh, Aditya and Goh, Gabriel and Agarwal, Sandhini and Sastry, Girish and Askell, Amanda and Mishkin, Pamela and Clark, Jack and others},
  booktitle={International conference on machine learning},
  pages={8748--8763},
  year={2021},
  organization={PMLR}
}

@inproceedings{paszke2019pytorch,
  title={{PyTorch:} An Imperative Style, High-Performance Deep Learning Library},
  author={Adam Paszke and Sam Gross and Francisco Massa and Adam Lerer and James Bradbury and Gregory Chanan and Trevor Killeen and Zeming Lin and Natalia Gimelshein and Luca Antiga and Alban Desmaison and Andreas Köpf and Edward Yang and Zach DeVito and Martin Raison and Alykhan Tejani and Sasank Chilamkurthy and Benoit Steiner and Lu Fang and Junjie Bai and Soumith Chintala},
  booktitle={Proceedings of the 33rd International Conference on Neural Information Processing Systems (NeurIPS)},
  pages={8026--8037},
  year={2019}
}

@inproceedings{liu2023g,
  title={{G-Eval: NLG Evaluation using GPT-4 with Better Human Alignment}},
  author={Liu, Yang and Iter, Dan and Xu, Yichong and Wang, Shuohang and Xu, Ruochen and Zhu, Chenguang},
  booktitle={EMNLP},
  pages={2511--2522},
  year={2023}
}

@inproceedings{zhangbertscore,
  title={BERTScore: Evaluating Text Generation with BERT},
  author={Zhang, Tianyi and Kishore, Varsha and Wu, Felix and Weinberger, Kilian Q and Artzi, Yoav},
  booktitle={International Conference on Learning Representations},
  year={2020}
}

@misc{honnibal2017spacy,
  title={spaCy 2: Natural language understanding with Bloom embeddings, convolutional neural networks and incremental parsing},
  author={Honnibal, Matthew},
  year={2017}
}

@inproceedings{vedantam2015cider,
  title={{CIDEr:} Consensus-based Image Description Evaluation},
  author={Vedantam, Ramakrishna and Lawrence Zitnick, C and Parikh, Devi},
  booktitle={Proceedings of the IEEE Conference on Computer Vision and Pattern Recognition (CVPR)},
  pages={4566--4575},
  year={2015}
}

@inproceedings{denkowski2014meteor,
  title={Meteor Universal: Language Specific Translation Evaluation for Any Target Language},
  author={Denkowski, Michael and Lavie, Alon},
  booktitle={Proceedings of the Ninth Workshop on Statistical Machine Translation},
  pages={376--380},
  year={2014}
}

@inproceedings{lin2004rouge,
  title={Rouge: A package for automatic evaluation of summaries},
  author={Lin, Chin-Yew},
  booktitle={Text summarization branches out},
  pages={74--81},
  year={2004}
}

@inproceedings{papineni2002bleu,
  title={BLEU: a method for automatic evaluation of machine translation},
  author={Papineni, Kishore and Roukos, Salim and Ward, Todd and Zhu, Wei-Jing},
  booktitle={Proceedings of the 40th Annual Meeting on Association for Computational Linguistics},
  pages={311--318},
  year={2002}
}
